\documentclass[doc,floatsintext]{apa6}\usepackage{knitr}  
\usepackage[german,american]{babel}
\usepackage[utf8]{inputenc}
\usepackage{pdflscape} 
\usepackage{rotating}
\usepackage{csquotes}
\usepackage{longtable}
\usepackage{float}
\usepackage{amsmath}
\usepackage{calrsfs}
\usepackage{amssymb}
\usepackage{multirow}
\usepackage{multicol}
\usepackage{fancyvrb}
\usepackage{listings}
\usepackage{courier}
\usepackage{setspace}
\usepackage{cases}
\usepackage{todonotes}
 \usepackage{relsize}

\newsavebox{\LstBox}

\usepackage{xcolor}
        
\usepackage{apacite}
\usepackage{tikz-cd}
\usetikzlibrary{arrows,shapes,positioning,shadows,trees,matrix}

\usepackage{forest}
\usepackage{enumerate}
\usepackage{color}
\usepackage{gb4e}
\noautomath
\usepackage[all]{xy}
\xyoption{line}

\newcommand{\posstextcite}[1]{\citeauthor{#1}'s \citeyear{#1}}

\title{Models of retrieval in sentence comprehension: A computational evaluation using Bayesian hierarchical modeling}
\twoauthors{Bruno Nicenboim}{Shravan Vasishth}
\twoaffiliations{Department of Linguistics, University of Potsdam,\\ Potsdam, Germany\\E-mail:bruno.nicenboim@uni-potsdam.de}{Department of Linguistics, University of Potsdam, Potsdam, Germany}
\authornote{The work was supported by  Potsdam Graduate School, and the University of Potsdam. }
\shorttitle{Models of Retrieval}
\leftheader{Models of Retrieval}

\abstract{
Research on similarity-based interference has provided extensive
evidence that  the formation of dependencies between non-adjacent words relies
on a cue-based retrieval mechanism. There are two different models that can
account for one of the main predictions of interference, i.e., a slowdown at a
retrieval site, when several items share a feature associated with a retrieval
cue: \posstextcite{LewisVasishth2005} activation-based model and
\posstextcite{McElree2000} direct-access  model. Even though these two models
have been used almost interchangeably, they are based on different assumptions
and predict differences in the relationship between reading times and response
accuracy. The activation-based model follows the assumptions of the ACT-R
framework, and its retrieval process behaves as a lognormal race between
accumulators of evidence with a single variance. Under this model, accuracy of
the retrieval is determined by the winner of the race and retrieval time by
its rate of accumulation. In contrast, the direct-access model assumes a model
of memory where only the probability of retrieval can be affected, while the
retrieval time is drawn from the same distribution; in this model, differences in latencies are a
by-product of the possibility of backtracking and repairing incorrect
retrievals. We implemented both models in a  Bayesian hierarchical framework
in order to evaluate them and compare them. { {The
data show that correct retrievals take longer than incorrect ones, and this   
pattern is better fit  under the direct-access model than under the
activation-based model.}} This finding does not rule out the
possibility that retrieval may be behaving as a race model with assumptions
that follow less closely the ones from the ACT-R framework. By
introducing a modification of the activation model, i.e, by assuming that the
accumulation of evidence for retrieval of incorrect items is not only slower
but noisier (i.e., different variances for the correct and incorrect items),
the model can provide a fit as good as the one of the direct-access model. 
{ This first ever computational evaluation of alternative accounts
of retrieval processes in sentence processing opens the way for a broader investigation of theories
of dependency completion.}
}

\note{Draft of \today.}


\authornote{We thank Hedderik van Rijn for the useful discussion during
Groningen Spring School on Cognitive Modeling, 2016, and  Pavel Loga{\v c}ev
for insightful comments on the manuscript. This work was supported by Potsdam
Graduate School, the University of Potsdam, and by Volkswagen Foundation grant
89953.}
\IfFileExists{upquote.sty}{\usepackage{upquote}}{}
\begin{document}
\maketitle

There is strong evidence that the formation of syntactic dependencies between
non-adjacent words relies on the memory system. An example is the so-called
locality effect: increasing the distance between co-dependents (such as
subjects and verbs)  tends to lead to greater processing difficulty
\cite{Gibson2000,GrodnerGibson2005}. Research on interference makes a
similar point: the speed and/or accuracy of dependency completion can be
adversely affected by the presence of items in memory that are similar to the
retrieval target  \cite<among
others:>[]{GordonEtAl2002,VanDykeLewis2003,vanDyke2007,JaegerEngelmannVasishth2015,NicenboimEtAl2016NIG,VasishthEtAl2008}.  Such a central role for
memory in sentence comprehension is well-motivated: it is implausible that the
parser could keep track of a large and in principle unbounded inventory of the
dependencies that can be found in a sentence, since they easily  exceed the
amount of information that can be held in the focus of attention 
\cite{McElreeDosher1989,McElree2006,Cowan1995,Oberauer2013,Marcus2013}.  The
evidence from studies investigating similarity-based interference
\cite<see the meta-analysis of published reading studies in>[]{JaegerEtAl2017}
suggests that dependency completion relies on a \emph{content-addressable 
cue-based retrieval mechanism} that is subject to interference
\cite{McElree2000,VanDykeLewis2003,LewisEtAl2006}. Similarity-based
interference is a phenomenon that is not unique to language, and occurs when
several items share a feature associated with a retrieval cue. A major
implication is that the retrieval mechanism employed for the creation of
linguistic dependencies is similar to the one utilized in non-language
domains.

There are multiple implementations compatible with such a content-addressable
cue-based retrieval mechanism in sentence processing. As a verbally stated
model,  this type of mechanism would entail that when retrieval cues fully
match the target of retrieval, similarity-based interference would cause an
inhibitory effect, that is, an increase of processing difficulty at the
retrieval of a dependent. This processing difficulty would be reflected in
longer reading times and lower accuracy. However, in some cases, shorter
reading times have been observed when  increased processing difficulty was
clearly expected \cite{VanDykeMcElree2006,NicenboimEtAl2015Frontiers,NicenboimEtAl2016Frontiersb}. In these cases, it is usually assumed that the
fast reading times are a consequence of a shallow parse \cite<due to, for
example, good-enough processing,>[]{FerreiraEtAl2002} caused by cognitive
overload. There can be good reasons to assume that shorter reading times are
associated with increased difficulty, for example, when shorter reading times
co-occur with  lower comprehension accuracy \cite{VanDykeMcElree2006} or
lower working memory capacity
\cite{NicenboimEtAl2015Frontiers,NicenboimEtAl2016Frontiersb}. However,
the trade-off between reading times and comprehension accuracy is usually left
underspecified.

There are two models that make explicit the relationship between reading times
and retrieval accuracy, and even though they are sometimes not differentiated,
they constitute two different implementations of the content-addressable
cue-based retrieval mechanism. These model are the
\posstextcite{LewisVasishth2005} \emph{activation-based} model, and
\posstextcite{McElree2000}  \emph{direct-access} model. These models have
different implications for retrieval processes in sentence comprehension. The
activation-based model assumes a process that resembles a race model
\cite<>[]{AudleyPike1965,Vickers1970}, where evidence for each
retrieval candidate is accumulated with different rates. This race determines
both the latencies and the retrieval accuracy. By contrast, the direct-access
model assumes that retrieval candidates have different levels of
\emph{availability}, which is the probability that a memory representation is
retained. Availability determines only the accuracy of the retrieval and not
the latency. In this model, a difference in latency between two conditions is
a by-product of a mixture of directly accessed items, and retrievals that were
initially incorrect, but they are reanalyzed leading to a correct retrieval.

The goal of this paper is to unpack the quantitative predictions of the
activation-based and direct-access models by implementing them in a  Bayesian
hierarchical framework. This will allow us to compare their relative fit to a
representative dataset and to assess their validity as models of retrieval
that can account for similarity-based interference. We used a subset of the
data from \citeA{NicenboimEtAl2016NIG}, which investigated similarity-based
interference from the number feature using two relatively large-sample
self-paced reading experiments. The data in this study include two dependent
measures: (i) reading times for the critical region where retrieval from
memory is assumed to occur, and (ii) accuracies in a comprehension task that
targets specific dependency relations through a multiple choice task. This
dataset is especially suitable for our modeling purposes because, apart from
\citeA{vanDyke2007},who also evaluated some of the dependencies, this is
the only dataset that we are aware of that uses comprehension questions to
directly assess the resolution of the dependencies. As explained in detail
later, these two dependent measures (reading times and accuracy) are necessary
for evaluating the models. We begin by describing the \citeauthor{NicenboimEtAl2016NIG} study.

\subsection*{The \citeauthor{NicenboimEtAl2016NIG} study}
\citeA{NicenboimEtAl2016NIG} used stimuli like \eqref{ex:exp1}. There were
two conditions, high vs.\ low interference, which were assumed to affect the
dependency between the subject (i.e., \textit{Der Wohltäter} ``The
philanthropist'') and the verb (i.e., \textit{begrüßt hatte} ``had greeted'').
In the high interference condition, two nouns intervened between these two
co-dependents that had the same number marking as the target noun, the subject
of the sentence, namely, singular marking. In the low interference case, the
two intervening nouns had plural marking while the target noun remained
singular. In German, the verb
(i.e.,\textit{begrüßt hatte})  agrees in number with its subject; in the
high interference condition, the retrieval cue set at the verb to seek out a
singular noun would match three nouns. By contrast, in the low interference
condition, only one noun matches this retrieval cue. Thus, reading time at the
critical region, the verb
\textit{begrüßt hatte}, provides an estimate of any interference effect.

Each target sentence was followed by a question that queried either the subject of the
matrix verb (e.g., ``sat''), the subject of the embedded verb (e.g.,``had
greeted''), or the object of the embedded verb. The possible answers were
provided in multiple-choice format in pseudo-randomized order. For all the questions,
participants had the option to answer ``I don't know'', when they did not
remember or could not answer.

\begin{exe}
    \ex  \label{ex:exp1}
    \begin{xlist}
        \ex \textsc{High Interference} \label{ex:hard}
        \gll \textbf{Der} \textbf{Wohltäter}, der den Assistenten {} des
        Direktors \textbf{begrüßt} \textbf{hatte}, saß später im
        Spendenausschuss.\\
        \textbf{The.sg.nom} \textbf{philanthropist}, who.sg.nom
        the.\underline{sg}.acc assistant (of) the.\underline{sg}.gen director
        \textbf{greeted} \textbf{had.sg}, sat.sg {later} {in the} {donations
        committee}.\\
        \glt ‘The philanthropist, who had greeted the assistant of the director,
        sat later in the donations committee.'
        \ex \textsc{Low Interference} \label{ex:easy}
        \gll \textbf{Der} \textbf{Wohltäter}, der die Assistenten {} der
        Direktoren  \textbf{begrüßt} \textbf{hatte}, saß später im
        Spendenausschuss.\\
        \textbf{The.sg.nom} \textbf{philanthropist}, who.sg.nom
        the.\underline{pl}.acc assistants (of) the.\underline{pl}.gen
        directors \textbf{greeted} \textbf{had.sg}, sat.sg {later} {in the}
        {donations committee}.\\
        \glt ‘The philanthropist, who had greeted the assistants of the
        directors, sat later in the donations committee.'
    \end{xlist}
\end{exe}

\begin{figure}[h]
\begin{knitrout}
\definecolor{shadecolor}{rgb}{0.969, 0.969, 0.969}\color{fgcolor}

{\centering \includegraphics[width=.5\linewidth]{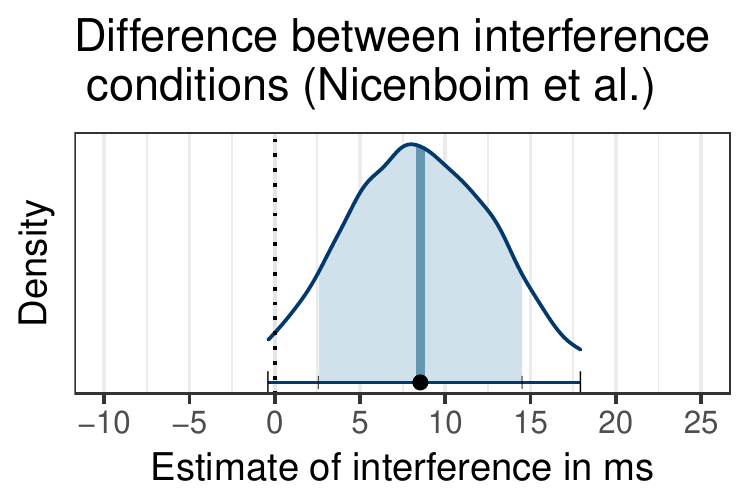} 

}

\end{knitrout}
\caption{Posterior distribution of the estimated difference in reading times between
conditions (in milliseconds)  for the pooled data of Nicenboim et al. The vertical line indicates the mean of the
posteriors, the outer error bars demarcate the 95\% credible intervals, and
the inner error bars and filled section of the distributions the 80\% credible
interval.}\label{fig:NIG-ms}
\end{figure}

\citeA{NicenboimEtAl2016NIG} { ran two experiments that showed}
an inhibitory effect of similarity-based interference for the retrieval of the
subject (``the philanthropist'') when it shared the number feature singular
with other competitor NPs (``the assistant'', ``the director''), that is
retrieval in high interference conditions took longer than in low interference
conditions. Given that the auxiliary verb (\textit{hatte} ``had'') is
morphologically marked as singular, the longer reading times at the auxiliary
verb is consistent with cue-based retrieval. { Figure
\ref{fig:NIG-ms} shows the posterior distribution of the difference of reading
times between conditions. }
 
Both the activation-based  and the direct-access models make
the correct predictions regarding the average behavior---both predict an inhibitory
interference effect. However, these two accounts differ in the way that 
correct and incorrect retrievals occur, and these different underlying mechanisms
can be investigated using the data from \citeA{NicenboimEtAl2016NIG}.

We first describe the models qualitatively, but in order to unpack the
predictions of these two models we later provide a more formal presentation.
We then evaluate the models quantitatively by examining the relationship
between reading times and accuracy.

There are two main findings in the present study: First, the direct-access
model provides a better fit to the data than the activation-based model.
Second, we show that a variation of the activation-based model fits the data
as well as the direct-access model, and also provides a reasonable model of
the underlying generative process. { {A
surprising aspect of the model selection is that it is driven by the
difference in latencies for correct and incorrect retrievals, and not by the
difference between interference conditions.}}\label{p:surprising}

\section{Overview of the activation-based and direct-access model}

\paragraph{The activation-based model} 

The activation-based model as implemented by \citeA{LewisVasishth2005} is a
computational model of sentence processing in which dependencies of non-adjacent 
elements are created via a content-addressable cue-based retrieval
mechanism. This model was realized in ACT-R \cite{AndersonEtAl2004},
which is a general cognitive architecture used to model a vast variety of
cognitive phenomena. { The activation-based model uses the
declarative retrieval module of ACT-R, which has been shown to be able to
account for many memory-related  phenomena \cite<e.g.,>[]
{AndersonEtAl1998,AndersonReder1999,VanRijnAnderson2003}. } This means  that
sentence processing depends on the application of general cognitive principles
to the specialized task of sentence parsing. Being a computational model, it
provides quantitative predictions of retrieval speed and accuracy.

The predictions regarding interference, locality, and some antilocality effects
of \posstextcite{LewisVasishth2005} original model  have been
investigated using simulations \cite{LewisVasishth2005,VasishthEtAl2008}. 
In addition, simplified versions 
of the model, which focused on certain aspects of the
architecture and evaluated some of the assumptions of the original model, have
also been used \cite<e.g.,>[]{DillonEtAl2013,DillonEtAl2014,KushEtAl2014,JaegerEngelmannVasishth2015,VasishthLewis2006,NicenboimEtAl2016Frontiersb,Engelmann2015,Parker2016,Parker2017,EngelmannJaegerVasishth2017}.

Crucially, the activation-based model provides an account of the relationship
between reading times at the dependency resolution site and the accuracy of
the resolution. This is so because dependency creation relies on the retrieval
of the correct item from memory; in ACT-R terms, what is retrieved is a chunk.
The chunk with the highest activation is retrieved and its activation level
determines the retrieval time { with higher activation leading to shorter times}. 
We provide next an informal explanation of the
key aspects of the activation-based model. 
We do this using example \eqref{ex:exp1} from \citeA{NicenboimEtAl2016NIG}. We show that the
activation-based model can explain similarity-based interference
effects, predicting inhibitory interference (i.e., an increase in processing
difficulty) when a competitor NP matches the singular number feature of the
target of retrieval.

The main assumptions of the model are that (i) words and phrases are encoded
in memory as bundles of features \cite<as
in>[]{Nairne1990,OberauerKliegl2006} that include lexical, semantic, and
syntactic information, and that (ii) retrieval cues are used to identify the
``correct'' chunk from memory: If retrieval cues (which are feature
specifications) match with the features of a chunk in memory, the chunk gets a
boost in activation, and if cues mismatch a chunk's features, activation is
decreased. Such a mechanism would always retrieve the correct item. However,
due to noise in the system, activation fluctuates randomly from trial to
trial, so that despite a cue match with a target, a competitor could have
higher activation and could end up being retrieved. An alternative possibility
is that all candidate chunks in memory fall below a retrieval activation
threshold (a parameter in ACT-R); in this case, retrieval would fail.

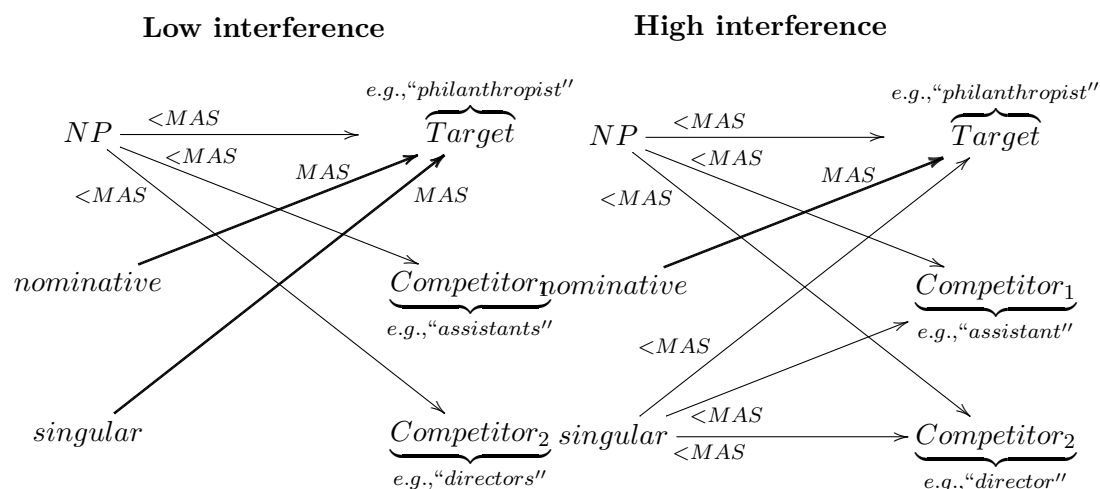
\begin{figure}
\begin{minipage}[c]{0.45\textwidth}
\centering
\textbf{Low interference}
\begin{displaymath}
\xymatrixcolsep{2.5cm}
\xymatrix{  
     NP \ar[r]^(0.25){<MAS}  \ar[dr]^(0.25){<MAS} \ar[ddr]_(0.15){<MAS}   &  {\overbrace{Target}^{e.g.,``philanthropist''}} \\
     nominative     \ar@<-0.1mm>[ur] \ar@<0.1mm>[ur] \ar[ur]^(.65){MAS} & {\underbrace{Competitor_1}_{e.g.,``assistants''}} \\
    singular  \ar@<-0.1mm>[uur] \ar@<0.1mm>[uur] \ar[uur]_(.85){MAS}   & {\underbrace{Competitor_2}_{e.g.,``directors''}}   }
\end{displaymath}
\end{minipage}
\begin{minipage}[c]{0.4\textwidth}
\centering
\textbf{High interference}
\begin{displaymath}
\xymatrixcolsep{2.5cm}
\xymatrix{  
     NP \ar[r]^(0.25){<MAS}  \ar[dr]^(0.25){<MAS} \ar[ddr]_(0.15){<MAS}   &  {\overbrace{Target}^{e.g.,``philanthropist''}} \\
     nominative     \ar@<-0.1mm>[ur] \ar@<0.1mm>[ur] \ar[ur]^(.65){MAS} & {\underbrace{Competitor_1}_{e.g.,``assistant''}} \\
    singular  \ar[uur]^(.25){<MAS} \ar[ur]_(.25){<MAS} \ar[r]_(.25){<MAS}   & {\underbrace{Competitor_2}_{e.g.,``director''}}   }
\end{displaymath}
\end{minipage}
\caption{Graph showing the associations between the cues NP, nominative, and
singular, and  the target and two competitor NPs. The width of
the arrow represents the strength of association, but this strength is
weighted, so two identical associations may assign different amount of
activation depending on the cue to which they belong. }
\label{fig:spread}
\end{figure}

As an example, consider the auxiliary verb (\emph{hatte}, ``had.sg'') of \eqref{ex:exp1}.
This is the region where an interference effect was seen in
\citeA{NicenboimEtAl2016NIG}. In both sentences \eqref{ex:hard} and
\eqref{ex:easy}, there is a dependency between this verb and its subject
(``the philanthropist''), and the only difference between the sentences is
that the intervening NPs (``the assistant/s'', ``the director/s'') appear in 
singular  in \eqref{ex:hard} and in plural  in \eqref{ex:easy}. The
activation-based model assumes that the feature information of each item such as
category, case, number, gender, and so forth  is encoded in memory. When the
embedded verb  (\emph{begrüßt hatte}, ``greeted had.sg'') is being read, 
an attempt is made to retrieve the subject. 
The verb
provides cues such as \emph{NP}, \emph{nominative} (notice that case
is encoded in the determiner of the NP in German),
\emph{singular}, among others features required from the target of the
retrieval. For each cue, a limited amount of activation (called the maximum
associated strength or $MAS$) is spread among the target and the competitors that are
stored in memory. 
The $MAS$ determines the strengths of association from each
cue to each item in memory. 
This strength of association represents how uniquely the cue
identifies a target. This means that in the low interference condition
\eqref{ex:easy}, the strength of association of the singular cue with the target is
determined by the maximum activation associated with this cue (since the cue
fully identifies the item). In the high interference condition
\eqref{ex:hard}, however, the target (``the philanthropist'') and the competitors
(``the assistant'', ``the director'') will be assigned some smaller part of
the maximum activation \cite<this is the so-called fan effect, for
details, see>[]{AndersonReder1999}, and thus their strength of association of
\emph{singular} will be smaller than the maximum activation. This is
regardless of the fact that in both conditions, there is another cue that
uniquely identifies the target, namely, being nominative: In both conditions
the target also receives activation due to the strength of association with
the cue nominative. This means that { (i) the target in high interference conditions
would receive (on average) less  total spreading activation than  the
target in low interference conditions, (ii)  the competitors in high
interference conditions would receive (on average) more  total spreading
activation than the competitors in low interference conditions, and (iii)
the target would receive (on average) more total spreading activation than the
competitors. See Figure \ref{fig:spread} for a schematic that illustrates this.
Given that higher activation leads to shorter retrieval times, the way
activation is shared among the NPs leads to inhibitory interference, that is,
longer retrieval times and lower accuracy in high interference conditions in
comparison with low interference ones. Another less studied consequence is that the
incorrect retrieval of the competitors should take longer than the correct
retrieval of the target. We will return to this last point in the discussion
section.

}

Given that both latency and probability of successful retrieval are affected by
activation, we will show later that the retrieval process is similar to a race
of accumulators \cite<among many others:>[]{AudleyPike1965,Vickers1970,UsherMcClelland2001}: Each item in memory is assigned an \emph{accumulator of
evidence} for its retrieval,  where the activation of each item acts as the
rate of accumulation. The accumulator that reaches the threshold of evidence
first determines which item is retrieved and with which latency. { This places the model
under a sequential sampling framework (\citeNP<such as the drift diffusion
model:>[]{Ratcliff1978,RatcliffMcKoon2008}; \citeNP<the leaky competitive
accumulator:>[]{UsherMcClelland2001}; \citeNP<linear deterministic models:>[among
others]{BrownHeathcote2008,HeathcoteLove2012}). } Furthermore,
some of the assumptions of ACT-R can allow us to frame the retrieval process
as one of simplest accumulator models: the lognormal race model with a single
variance for the noise associated with target, competitors, and failure accumulators
\cite{HeathcoteLove2012,RouderEtAl2014}. A variant that we will
consider later is a model with two separate variances, one for the target
accumulator, and one for the competitors and failure accumulators.

\paragraph{The direct-access model} 

It should be noted that a content-addressable system does not necessarily
entail a race between items in memory, and there are other models that are
also compatible with a content-addressable cue-based retrieval mechanism. The
cue-based retrieval model proposed in \citeA{VanDykeMcElree2006} is based
on McElree and colleagues' previous work
\cite<e.g.,>[]{McElree2000,McElreeEtAl2003} and, while it does not
assume a race model, it shares with the activation-based model some of the
assumptions of cue-based retrieval: Words and phrases are also encoded in
memory as feature bundles, and  retrieval cues are used to distinguish the
target from the competitors. { Whereas here the cues are combined
multiplicatively
\cite<as proposed in the Search of Associative Memory (SAM) model
of>[]{RaaijmakersShiffrin1980,GillundShiffrin1984}, \footnote{
\citeA{VanDykeMcElree2011} are explicit, however, in that it could also be
that cues are combined linearly, but with weights that are different enough so
that certain cues, such as syntactic cues, have a more prominent role. } } and
in the activation-based model the cues are combined additively, the result is
similar: The probability of retrieving a particular item from memory given the
retrieval cues is a function of the degree of the match between the cues and
the item, reduced by the degree to which the cues match other competitor items
in memory. However, in contrast with the activation-based account,  cues are
supposed to enable \emph{direct access} to relevant memory representations.
This means not only that there is no serial search between items in memory,
but that the distribution of access time is independent of the degree of match
between item and cue, and regardless of the quality or strength of the
representation of the item in memory \cite{McElree2000}. { As
a consequence of the direct access, in this model, the probabilities of
retrieval are affected by inhibitory interference, resulting in a pattern
similar to the one described for the activation-based model: (i) the
probability of retrieving the target in high interference conditions would be
lower than in low interference conditions, (ii) the probability of retrieving
one of the competitors in high interference conditions would be higher than in
low interference ones, and (iii) the probability of retrieving the target
would be higher than the probability of retrieving one of the competitors. }

It is not uncommon, however, that both poorer accuracy and longer reading
times associated with similarity-based interference are taken as evidence
for \posstextcite{McElree2000} direct-access model as well as for
\posstextcite{LewisVasishth2005} activation-based model. For example,
\citeA{VanDykeMcElree2006} write:
 
\begin{displayquote}
The current experiment supports a retrieval-based account of interference
effects in sentence processing, one that is compatible with the hypothesis
that a cue-based retrieval mechanism mediates the creation of grammatical
dependencies during parsing. One such mechanism has been proposed in Van Dyke
and Lewis (2003; see also Lewis and Vasishth, 2005; Van Dyke, 2002), in which
parsing success depends on the extent to which required constituents can be
retrieved from working memory. On this account grammatical heads provide
retrieval cues that are used to access previously stored items via a
content-addressable retrieval process (McElree, 2000, 2006; McElree, Foraker,
\& Dyer, 2003).
\end{displayquote}

A slowdown in self-paced reading or eyetracking-while-reading can also be
taken as evidence for direct access, since processing speed  may be affected
by differences in the likelihood of recovering an item from memory
\cite{McElree1993,McElreeEtAl2003}. This is because
\citeA{McElree1993} assumes that after a misretrieval, that is, an
incorrect or failed retrieval, the parser can often backtrack to reprocess the
retrieval and reach the appropriate analysis. This would mean that a correct
interpretation of a dependency could be arrived at because the correct
dependent was retrieved at the first attempt or, alternatively, because a
wrong dependent was retrieved initially but the parser backtracked and
retrieved the correct one. Given that  backtracking should take some
additional time, latencies associated with the correct responses would be a
mixture of fast directly accessed dependents and retrievals slowed down due to
the time needed for backtracking. Since interference adversely affects
retrieval probabilities, the proportion of errors would be higher in high
interference conditions. This would entail a higher proportion of backtracking
and hence slower latencies in the mixture of correct responses and on average,
high interference conditions would show longer reading times than low
interference conditions.

Since  the assumed constant distribution of retrieval times may not be
observable in reading for comprehension, evidence compatible with  the direct-access model but incompatible with the activation-based model comes only from
findings of speed-accuracy trade-off (SAT) procedure on rapid grammaticality
judgment task
\cite<e.g.>[]{McElree2000,McElreeEtAl2003,VanDykeMcElree2011}. In this
task, participants need to judge a sentence as either grammatical or
ungrammatical, and their judgment process is interrupted with a cue to respond
(typically a tone) after varying amounts of time
(\citeNP{Reed1973,Wickelgren1977}; \citeNP<and see also:>[]{ForakerMcElree2011}). 
However, this is a meta-linguistic task and it would be desirable if
independent support for the direct-access model could be found with
a reading-for-comprehension task. 
In addition, the conclusion that there is a constant distribution of retrieval times requires arguing
for a null result in one of the parameters of the SAT model. For SAT
procedures, accuracy is modeled as a function of three parameters
corresponding to the three phases of the SAT curve: (i) the asymptotic level
of performance, (ii) the intercept or the point in time where performance is
different from chance, and (iii) the rate at which accuracy grows from chance
to asymptote. While the presence of the effect in the asymptote, such that an
increase on interference lowers the asymptote, is evidence for the reduction
of the probability of accessing the target, the lack of evidence for changes
in the rate and intercept must be taken as evidence for no effect on the speed
of the retrieval. It could be, however, that the differences were too small to
be detected.

{{Since both models give virtually identical predictions
for non-SAT (self-paced reading or eye-tracking) experiments when comparing only
reading time averages across conditions, a slowdown at high
interference conditions has been taken as evidence for both models.}}
However, the two models assume  different relationships between retrieval
times and responses.  It is important to assess the fit to the data of each
model, since each one is compatible with different memory retrieval
mechanisms. 

\section{Implementation of the models}\label{sec:implementation}

In order to distinguish between the activation-based and direct-access models,
we implemented them as hierarchical Bayesian models. Implementing these models
affords several advantages: (i)  we can investigate (through posterior
predictive checking, explained below) whether the data could have been
generated by the models; (ii) we can determine how each model's parameters
were affected by interference effects; and (iii) the quality of fit of the two
models can be compared using cross-validation. {
{It is important to note that we are using Bayesian methods
as a flexible and interpretable way of extending models of cognitive processes
\cite{Lee2011,ShiffrinEtAl2008}, and this approach is orthogonal to the
question of whether the mind does or does not do Bayesian inference
\cite<see>[for a critical review on how probabilistic models of
perception and cognition should work]{Feldman2016}.} }

The benefits of using hierarchical Bayesian modeling are two-fold: (i) The
models incorporate the general advantages of Bayesian inference, such as the
use of credible intervals instead of confidence intervals, and the possibility
of fitting complex non-linear models \cite<see>[for an extended
discussion]{NicenboimVasishth2016}, and (ii) hierarchical models allow us to
take both between- and within-group variance into account and pool information
via shrinkage
\cite{GelmanEtAl2012}. This means that  we avoid overfitting the data and
at the same time we avoid aggregating data and losing valuable information about
group-level variability \cite{GelmanHill07}.  In the next section we provide a more
formal presentation of the assumptions and details of the implementation of
both models than the one given in the introductory overview.

\subsection{The activation-based model as a lognormal race}
In order to assess the fit of the activation-based account to the experimental
data, we implemented it as a lognormal race of accumulators with a shift
parameter and a single variance for the noise of all accumulators
\cite{RouderEtAl2014}, as explained further below. The retrieval in ACT-R
can be thought of as a decision processes, where target and competitors stored
in memory accumulate evidence until the first chunk reaches a certain
value and  is retrieved. Activation can be linked to evidence by assuming
that it represents the rate of its accumulation 
\cite<in a similar way as assumed by>[]{vanMaanenEtAl2011}. This is so because
activation in ACT-R represents the probability of the retrieval, and it is boosted by
processes that increase  evidence such as matching cues, previous
retrievals and so forth, while it is penalized by processes that decrease the
evidence such as mismatching features and decay.

{ Since the chunk with the highest activation is retrieved, the
equation that determines the latency of the retrieval in ACT-R
\eqref{eq:latency} ignores the activation of the other chunks. }Our
implementation, however, assumes that there is a \emph{potential} retrieval
time, or \emph{finishing time} in the race, $t_c$ for each candidate $c$ in
memory { that depends on its activation, $A_c$}. This  is the time
it would have taken for the chunk to be retrieved given its activation, as
shown in Equation \eqref{eq:tc}.

\begin{equation} \label{eq:latency}
Latency \propto  e^{-{\underset {c}{\operatorname {arg\,max} }}\,(A_c)}
\end{equation} 

\begin{equation} \label{eq:tc}
t_c \propto e^{-A_c}
\end{equation} 

The race model is implemented in the following way: Since the noise component
in $A_c$  is assumed to be normally distributed
\cite{LebiereEtAl1994} and affects the activations of all the chunks
to the same extent,\footnote{ The noise component is  sometimes approximated
to have a logistic distribution for convenience, see, for example
\citeA{Lebiere1999}. } for each trial, the finishing time of  each chunk,
$t_c$, is sampled from a lognormal distribution with the same  standard
deviation $\sigma$, and the fastest chunk in a given trial would be the one
retrieved (i.e., the chunk with the lowest $t_c$ in a given trial).
{ We account other aspects of processing (e.g., lexical access)
with a parameter $\gamma$. }

\begin{align}
t_c  \sim  e^{normal(- \mu_c + \gamma,\sigma)} \Rightarrow& \log(t_c)  \sim normal(- \mu_c+\gamma,\sigma)  \\
\Leftrightarrow& t_c \sim lognormal(- \mu_c +\gamma,\sigma) \label{eq:winner}
\end{align} 
where
\begin{align}
A_c &= \mu_c + \epsilon\\
\epsilon &\sim normal(0,\sigma)
\end{align}

The only observable data for every trial are (i)  the answer of the
comprehension question at the multiple choice task, $w$, (we assume that when
the question asks about the subject of the embedded verb, the answers
correspond to the chunk retrieved from memory, i.e., the winner of the race,
modulo offline distractions), and (ii) the reading times at the site of the
retrieval (the auxiliary verb ``had.sg''). The reading times will include the
retrieval time, $t_{c=w}$, of the ``winner'' chunk, and the time taken for
other processes. Given the evidence that distributions of  reaction times are
shifted
\cite{Rouder2005,NicenboimEtAl2016Frontiersb}, we assume a lower
bound, $\psi$, which represents changes in peripheral aspects of processing,
such as encoding or motor execution \cite{Rouder2005}. 

\begin{equation}
RT_{c=w} \sim  \psi + lognormal(- \mu_c  + \gamma,\sigma) \label{eq:RT1}
\end{equation} 
 
 Since the observed reading times, as shown in Equation \eqref{eq:RT1}, are
associated with the maximum activation (on a specific trial), the observable
data  constrain the unobserved  finishing times of all the other non-selected
choices $c$ which are not the winner; these finishing times must be slower than
the finishing time of the winner $w$. We can write this as Equation
\eqref{eq:forall}.

\begin{equation}
t_{\forall c, c\neq w} >  t_{c = w}  \label{eq:forall}
\end{equation}

Since we are not interested in the specific value of $\mu_c$, or $\gamma$, but in
learning from the model (i) whether the retrieval process resembles a race of
accumulators, and (ii) the effect of number interference on the target and
competitors, we fit the reading times, $RT$, as a function of $\alpha_c$
and an arbitrary constant, $b$, such that $b -
\alpha_c = -\mu_c+p$. By setting $b$ large enough (to 10,
for example), we ensure that $\alpha_c$ is strictly positive for ease of
interpretation: a higher positive number corresponds with a higher rate of
accumulation. \citeA{RouderEtAl2014} show that without further
assumptions,  thresholds and accumulation rates cannot be disentangled in the
lognormal race model. To estimate the thresholds in a lognormal race model,
\citeA{HeathcoteLove2012} assumed that both the rate of
accumulation $\nu$ and the thresholds $\eta$ were lognormally distributed, so
that the finishing times, $y$, were distributed in the following way $y \sim
lognormal(\mu= \mu_\eta - \mu_\nu, \sigma=\sqrt{
\sigma_\eta^2+\sigma_\nu^2})$. If the thresholds are fixed at some arbitrary
point $b$, then $\sigma=\sigma_\nu$ (since $\sigma_\eta=0$) and the rate of
accumulation $\mu_\nu = b- \mu$, thus we can interpret $\alpha_c$ as the rate
of accumulation associated with each chunk, and we can rewrite Equation \eqref{eq:RT1}
in the following way:

\begin{equation}
RT_{c=w} \sim  \psi + lognormal(b - \alpha_{c=w},\sigma) \label{eq:RT2}
\end{equation}

On every trial, $l$, we can estimate $\alpha_{l,c = w}$ from the observed
$RT$, and we can constrain the possible values of $\alpha_{l,\forall c, c\neq w}$ from
the values of $t_{l,\forall c, c\neq w}$ that could not be possible on a given trial.

Given that shifts  vary across participants but tend not to vary with
experimental manipulation \cite{Rouder2005}, we assume a certain shift
for every participant $i$, while $\alpha_{l,c}$, the activation together with
nuisance parameters, will vary by participant $i$ and by experimental item
$j$.

\begin{equation}
RT_{l,i,j}  \sim \psi_i +  lognormal(b - \alpha_{l,i,j,c=w} ,\sigma)
\end{equation}

In standard ACT-R, if the activation is below a certain threshold, $T$, the
retrieval fails with a latency that is proportional to $e^{-T}$ \cite{LebiereEtAl1994}.
To avoid a deterministic latency, we assign an accumulator to the possibility
of failure, which acts as a noisy timer, and its timeout depends on its
parameter $\alpha_{c = failure}$. In this way, the retrieval threshold can
be also thought as a chunk that competes for activation
\cite{VanRijnAnderson2003}.

\begin{figure}[p]
\begin{knitrout}
\definecolor{shadecolor}{rgb}{0.969, 0.969, 0.969}\color{fgcolor}

{\centering \includegraphics[width=.97\linewidth]{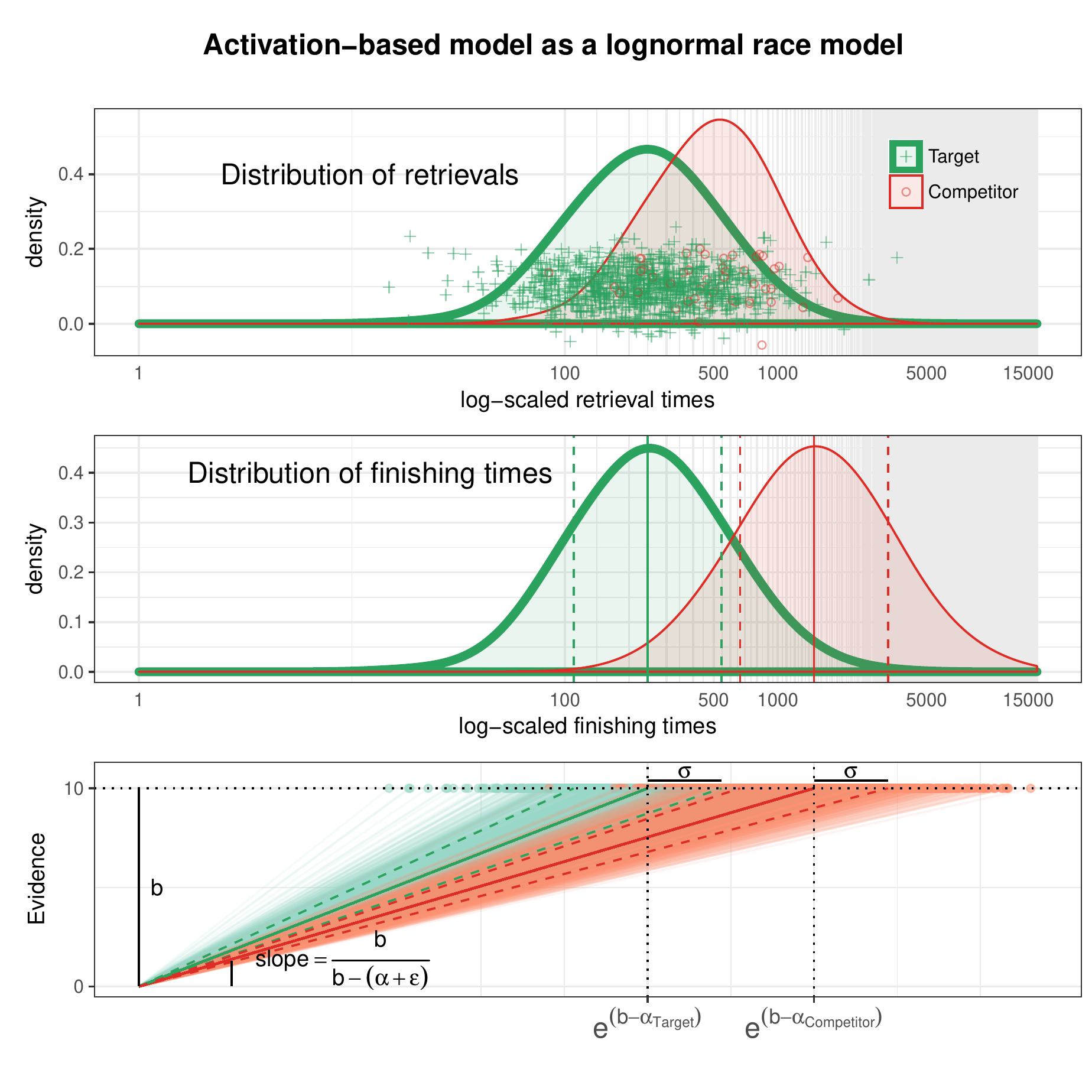} 

}

\end{knitrout}
\caption{The figure depicts how the distribution of retrievals is generated
from the activation-based model as a lognormal race model. The bottom figure
depicts the parameters of the activation-based model, the full lines in green
and red are the mean finishing times $t_{c=Target}$ and $t_{c=Competitor}$,
and the broken lines are finishing times one standard deviation away from the
mean. The middle figure shows the distributions of finishing times for target
and candidate; since every chunk is associated a potential finishing time,
$t$, both distributions have the same number of elements. The top-most figure
shows the distribution of retrieval times (adding the shift parameter $\psi$
would transform it to reading times); since only the winning chunks are
retrieved, the distribution of retrieval times for targets has more elements
than the one of the competitors. Notice that  the retrieval times are faster
than the finishing times: this is so because when a chunk has a very long
finishing time in a given trial, it is very likely that its competitor will be
faster. }\label{fig:race-ACTR}
\end{figure}

 Figure \ref{fig:race-ACTR}
summarizes the parameters and the process for two chunks. The lognormal race
model is sometimes called a \emph{ballistic} or \emph{deterministic} race
model because there is no within-choice noise (as in, for example, the drift
diffusion model): once a rate of accumulation is set for a given accumulator,
it will determine its time to the threshold; this is represented in the lower
part of Figure \ref{fig:race-ACTR} by the straight lines. This, however, does
not make the process deterministic, since the rate of accumulation changes
from trial to trial.

In the implementation of the activation-based model, the hierarchical
structure of this model is embedded in each parameter $\alpha_c$ associated
with each chunk $c$ (including one representing the failure) that is allowed
to vary by condition (high/low interference) and includes by-participants and
by-experimental-items intercepts and slopes (which are also allowed to be
correlated). This means that the model can account, for example, for an NP of
a certain experimental item being more semantically plausible as a retrieval
candidate than other NPs, by simply adjusting the by-item intercept of the
accumulators associated with each NP. See  Appendix \ref{app:ABM} for the
details of the Bayesian model.

\subsection{The direct-access model as a mixture model}

Next, 
we present an implementation of a direct-access model. This is a Bayesian hierarchical
implementation of the cue-based retrieval model proposed by
\citeA{VanDykeMcElree2006} based on McElree and colleagues' previous work
\cite<e.g.,>[]{McElree2000,McElreeEtAl2003}. Since the original model
has never been implemented computationally, it is underspecified in some
respects. We therefore made some assumptions in the model to 
spell these details out; these are described below.

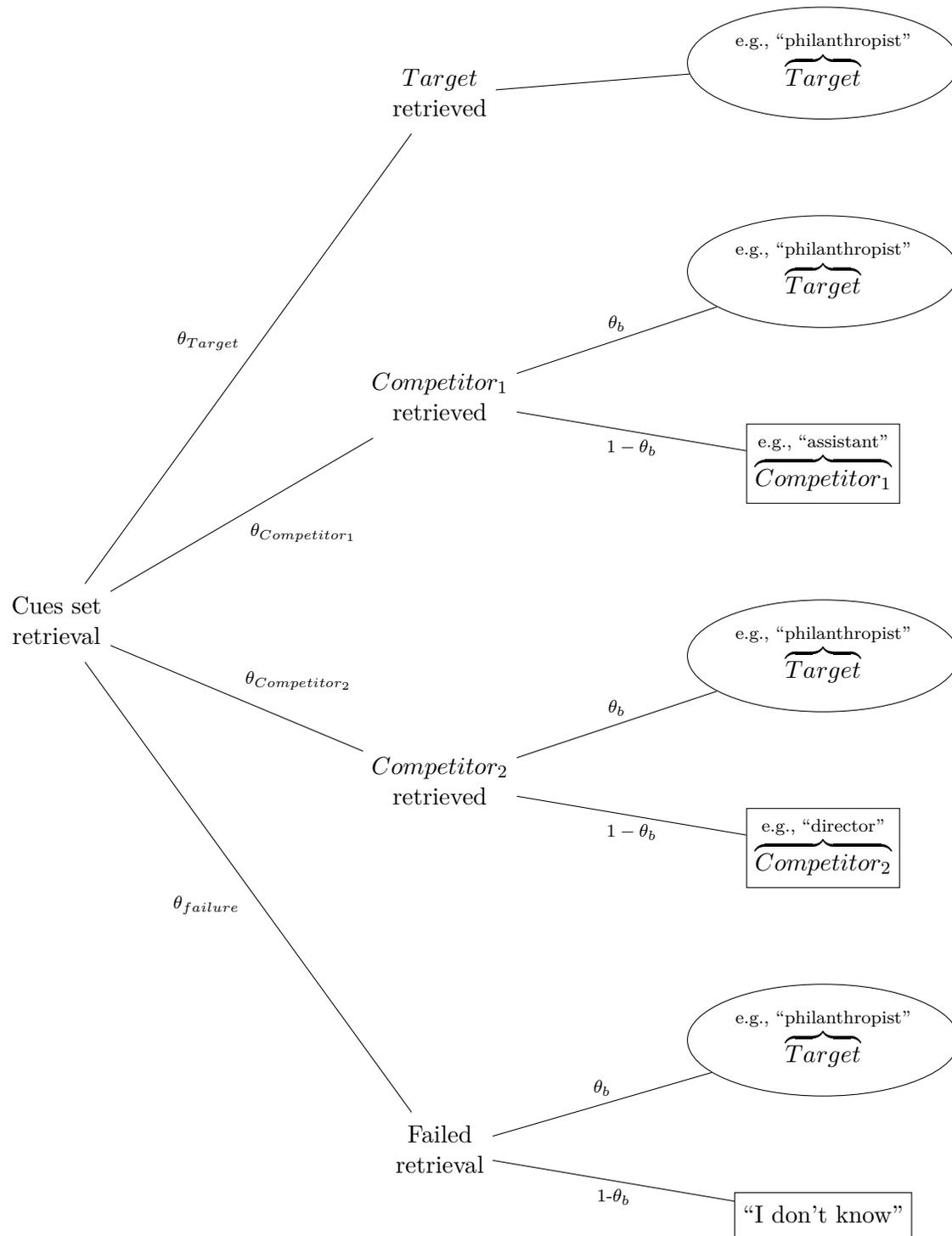
\begin{figure}[p]

\begin{forest}
where n children=0{tier=word}{}
  [Cues set\\ retrieval, for tree={grow=east,reversed,align=center,l=6cm,s sep=1.5cm} 
 [$Target$\\ retrieved,name=T, edge label={node[midway,font=\scriptsize,above left]{$\theta_{Target}$}} 
  [$\overbrace{Target}^{\text{e.g., ``philanthropist''}}$,ellipse,draw ]]
 [$Competitor_1$\\ retrieved,name=C1, edge label={node[midway,font=\scriptsize,below right]{$\theta_{Competitor_1}$}}
 [$\overbrace{Target}^{\text{e.g., ``philanthropist''}}$,ellipse,draw,edge label={node[midway,font=\scriptsize,above]{$\theta_b$}} ]
 [$\overbrace{Competitor_1}^{\text{e.g., ``assistant''}}$,draw,edge label={node[midway,font=\scriptsize,below]{$1-\theta_b$}} ]
 ] 
 [$Competitor_2$\\ retrieved,name=C2, edge label={node[midway,font=\scriptsize,above right]{$\theta_{Competitor_2}$}}
[$\overbrace{Target}^{\text{e.g., ``philanthropist''}}$,ellipse,draw, edge label={node[midway,font=\scriptsize,above]{$\theta_b$}} ] 
[$\overbrace{Competitor_2}^{\text{e.g., ``director''}}$,draw, edge label={node[midway,font=\scriptsize,below]{$1-\theta_b$}} ]
 ]
[Failed\\ retrieval,name=F, edge label={node[midway,font=\scriptsize,below left]{$\theta_{failure}$}}
[$\overbrace{Target}^{\text{e.g., ``philanthropist''}}$,ellipse,draw,edge label={node[midway,font=\scriptsize,above]{$\theta_b$}} ]
[ ``I don't know'',draw,edge label={node[midway,font=\scriptsize,below]{1-$\theta_b$}} ]
 ]
]
\end{forest}
\caption{Graph showing the different responses as a categorical distribution with correct responses inflated by the probability of backtracking, $\theta_b$. Correct responses are inside ellipsis and incorrect ones inside rectangles.}
\label{fig:daprob2}
\end{figure}

In order to account for differences in reading times, the direct-access model
assumes that, in some proportion of the cases, the parser is able to backtrack
a misretrieval and to access the target candidate, taking some extra time.
Thus the reading times associated with the correct retrievals are a mixture of
directly accessed as well as backtracked retrievals as shown in Figure
\ref{fig:daprob2}. This means that the probability of the final retrieval of a
certain chunk, $P_r$ (which should be equivalent to the proportion of
responses given at the multiple choice task modulo offline distractions),
{ is affected by the probability of backtracking, $\theta_b$,
together with the probability of either an initial retrieval of the target,
$\theta_{Target}$, an initial retrieval of one of the competitors,
$\theta_{Competitor_{1,2}}$, or an initial failure, $\theta_{failure}$, in the
following way:

\begin{align}
P_r(Target) &= \theta_{Target} + (1-\theta_{Target}) \cdot \theta_b\\
P_r(Competitor_{\{1,2\}}) &= \theta_{Competitor_{\{1,2\}}}   \cdot (1 - \theta_b) \\
P_{failure} &= \theta_{failure}  \cdot (1 - \theta_b)
\end{align} 

\noindent
where

\begin{equation}
\theta_{Target} + \theta_{Competitor_1} + \theta_{Competitor_2} + \theta_{failure} =1
\end{equation}
}

The core assumption of the direct-access model is that retrieval  takes the
same time on average, $t_{da}$, regardless of the availability of the
to-be-retrieved chunk. This is in contrast with the activation-based model, but also in
contrast with the SAM framework of \cite{GillundShiffrin1984}, where the
retrieval time depends on the match between cues and features of the chunk.
The implications for the direct-access model is that there will be two
different distributions of reading times: a distribution associated with the
incorrect responses and a distribution associated with the correct ones. An
incorrect response is given by a participant, only when the wrong chunk is
retrieved and there is no backtracking and repair process; see Figure
\ref{fig:daprob2}. In this case, the reading times at the retrieval site will
only include the time needed for the direct access, $t_{da}$, together with
the time taken for other processes, $\gamma$, and normally distributed noise with
standard deviation $\sigma$; crucially, this time is independent of the
level of interference. We assume, as before, that the noise and the time taken
for other processes are added to the location of the lognormal distribution;
in addition,  reading times are assumed to be shifted by some minimum amount
$\psi$ of time that represents the lower bound of the process. Thus, we can
assume that the reading times at the retrieval site for each trial $l$ that
are associated with an incorrect response (i.e., the retrieval of a competitor
or a failed retrieval) have a shifted lognormal distribution, where the
$t_{da}$ depends on each participant $i$, and experimental item $j$, but not
on the experimental condition or the identity of the chunk retrieved.

\begin{equation}
RT_{incorrect,l,i,j}  \sim \psi_i +  lognormal(t_{da,i,j}+\gamma_{i,j},\sigma)
\end{equation} 

For  correct responses, the reading times depend on whether there is a repair
process or not; see Figure \ref{fig:daprob2}. This entails that the
distribution of reading times  is  a mixture of two components: The first one
is associated with chunks correctly retrieved at the first attempt as shown in the first line of
Equation~\eqref{eq:cases}, and it is identical to the distribution of incorrect
responses. The second one is associated with incorrect responses that are
backtracked and repaired, and it includes the direct-access times, $t_{da}$,
together with the time it takes to backtrack and do a reanalysis, $t_{b}$, as
shown in the second line of Equation \eqref{eq:cases}. (We derive the exact proportions in Appendix
\ref{app:DAM}.)

\begin{equation}
RT_{correct,l,i,j}  \sim  \psi_i +
\begin{cases}
lognormal(t_{da,i,j}+\gamma_{i,j},\sigma) & \text{, if the first try is correct } \label{eq:cases} \\
lognormal( t_{da,i,j}+t_{b,i,j}+\gamma_{i,j},\sigma) & \text{, otherwise} 
\end{cases}
\end{equation} 

As with the activation-based model, we are not interested in $\gamma$, and we
define $T_{da,i,j} =t_{da,i,j} +\gamma_{i,j} $. 
In the implementation of the model, we estimate the probability of each retrieval and
the effect of interference on the retrieval probability using a multi-logit
regression (or categorical distribution with the parameters on the logit
scale). This is achieved by  assigning a hierarchical structure to the
parameters of the multi-logit regression which vary by condition (high/low
interference) and include by-participants and by-experimental-items intercepts
and slopes which have one correlation matrix for participants and one matrix
for experimental items. Furthermore, we assign a hierarchical structure also
to $T_{da}$ and $t_{b}$, which are composed by by-participant and
by-experimental-item varying intercepts. To allow for correlations between the
direct access time and the backtracking time, we included also one correlation
matrix for participants and one matrix for experimental items. This means that
we assume that latencies should not be affected by retrieval probabilities.
See Appendix
\ref{app:DAM} for more details.


\section{Evaluation of the activation-based and direct-access models}

\subsection{Application to data from a self-paced reading experiment} \label{ss:spr}

{
{
We fit the models to a subset of the data from \citeA{NicenboimEtAl2016NIG}.
This work reports two self-paced reading studies: an exploratory and a
confirmatory study  with a total of 183 participants investigating
similarity-based interference. Both studies included the same 60 experimental
items, where a high interference condition with two competitor nouns sharing the
singular feature with the target of the retrieval (e.g., ``the director'') is
contrasted with a low interference condition with two plural competitors, see
\eqref{ex:exp1}, repeated here for convenience as
\eqref{ex:exp1a}. In order to encourage attentive reading, after every sentence
participants answered a multiple choice questions querying different aspects of
the sentence: for example, \eqref{ex:multip-q}.

\begin{exe}
    \ex  \label{ex:exp1a}
    \begin{xlist}
        \ex \textsc{High Interference} \label{ex:harda}
        \gll \textbf{Der} \textbf{Wohltäter}, der den Assistenten {} des
        Direktors \textbf{begrüßt} \textbf{hatte}, saß später im
        Spendenausschuss.\\
        \textbf{The.sg.nom} \textbf{philanthropist}, who.sg.nom
        the.\underline{sg}.acc assistant (of) the.\underline{sg}.gen director
        \textbf{greeted} \textbf{had.sg}, sat.sg {later} {in the} {donations
        committee}.\\
        \glt ‘The philanthropist, who had greeted the assistant of the director,
        sat later in the donations committee.'
        \ex \textsc{Low Interference} \label{ex:easya}
        \gll \textbf{Der} \textbf{Wohltäter}, der die Assistenten {} der
        Direktoren  \textbf{begrüßt} \textbf{hatte}, saß später im
        Spendenausschuss.\\
        \textbf{The.sg.nom} \textbf{philanthropist}, who.sg.nom
        the.\underline{pl}.acc assistants (of) the.\underline{pl}.gen
        directors \textbf{greeted} \textbf{had.sg}, sat.sg {later} {in the}
        {donations committee}.\\
        \glt ‘The philanthropist, who had greeted the assistants of the
        directors, sat later in the donations committee.'
    \end{xlist}
\end{exe}

\begin{exe}
  \ex   Wer hatte jemanden begrüßt?  \label{ex:multip-q}
  \glt Who had greeted someone?
  \begin{xlist}
  \begin{multicols}{2}
    \ex  der/die Wohltäter \textsc{(correct)}
    \glt the philanthropist(s) 
    \ex  der/die Assistent/en 
    \glt the assistant(s) 
\columnbreak
    \ex  der/die Direktor/en
    \glt the director(s)
    \ex  Ich weiß es nicht
    \glt I don't know
\end{multicols}

  \end{xlist}
\end{exe}

For the current study, we pooled the data of both self-paced reading
experiments,  keeping only the sentences with questions that queried the subject
of the embedded verb, as in \eqref{ex:multip-q}. This was done because cue-based
retrieval models predict that interference will affect only retrievals where the
cue is relevant. This left us with 20 sentences for each participant. For each
sentence we used  two dependent variables: (i) the time taken for reading the
auxiliary verb \emph{hatte} (``had.sg''), where \citeA{NicenboimEtAl2016NIG}
found evidence for interference effects in reading times, and (ii) the response
given at the multiple choice task.

In order to fit the models, however, we need \emph{retrieval times} and
the \emph{retrieved argument}. Thus we assume that all things being equal (as in an
experimental setting), (i) reading times at the auxiliary verb include the
retrieval time (as well as other theoretically uninteresting processes), and
(ii) participants' answers are informative of the dependency completion they
carried out.
}}

We fit the models using \emph{rstan} package \cite<>[version 2.15.1]{RStan2017} in R
\cite{R2015} with four chains and 3000 iterations, half of which were
the burn-in or warm-up phase. In order to assess convergence, we verified that
the potential scale reduction factors, $\hat{R}$s, of the parameters were close
to one, and we also visually inspected the chains \cite<>[section
11.4]{GelmanEtAl2014}. This indicates that the chains have mixed and they are
transversing the same distribution.  When needed, we also increased the maximum
tree-depth and the adaptation parameter $\delta$ of the sampler to eliminate
divergent transition and achieve convergence. We also verified that we could
recover the parameters from the models using fake-data simulation
\cite{GelmanHill07}, see Appendix~\ref{app:rec}.\footnote{Data and code can
be downloaded from:
\url{www.ling.uni-potsdam.de/~nicenboim/code/code-data-retrieval-models.zip}}

\subsection{Posterior predictive checking} 
We use posterior predictive checking to  examine the descriptive adequacy of
the models (\citeNP{ShiffrinEtAl2008}; \citeNP<>[Chapter 6]{GelmanEtAl2014}), that
is, the observed data should look plausible under the posterior predictive
distribution. The posterior predictive distribution is composed of 6000
datasets (one for each iteration) that the model generates based on the
posterior distributions of its parameters. In other words, given the posterior
of the parameters of the model (which are based on the current data), the
posterior predictive distribution shows how other data may look like.
Achieving descriptive adequacy means that the current data could have been
predicted with the model. While passing a test of descriptive adequacy is not
strong evidence in favor of a model, a major failure in descriptive adequacy
can be interpreted as strong evidence against a model
\cite{ShiffrinEtAl2008}. Thus, posterior predictive checking is an
important sanity check to assess whether the model behavior is reasonable
\cite<see>[for further discussion]{GelmanEtAl2014}

Given that the main difference between the activation-based model and the
direct-access model is in the way they account for the relationship between
retrieval probability and latencies, for each of the 6000 datasets generated
by the models, we calculate the means and $.1$--$.9$ quantiles of the reading times
associated with each response, as well as the mean proportion of responses
given. We represent this graphically using violin plots
\cite{HintzeNelson1998}: the width of the violin plots represents the
density of the predicted means (or quantiles). The observed mean (or quantile)
of the data is represented with a cross. If the data could  plausibly have
been generated by the model, we would expect the crosses to be inside the
violin plots.

\subsection{Estimation of relevant parameters} 
In addition to fitting the data, the models include parameters that can be
interpreted and can give support (or falsify) some assumptions of the effect of
interference under the two presented models. { In particular, the
following inequalities represent the expected relationships between activations
for the activation-based model and how they are affected by interference (with
$|HI$ indicating under high interference, and $|LI$ under low interference):

\begin{align}   
{\alpha}_{Target|HI} & <  {\alpha}_{Target|LI}            \label{eq:a-tt} \\
{\alpha}_{Target} & >  \{{\alpha}_{Competitor_1},{\alpha}_{Competitor_2}, \alpha_{Failure}\}   \label{eq:a-tc}\\
{\alpha}_{Competitor_{\{1,2\}}|HI} & >  {\theta}_{Competitor_{\{1,2\}}|LI} \label{eq:a-cc}
\end{align}

\noindent
and similarly for the probability of retrieval at the first attempt in the direct-access model:

\begin{align}
{\theta}_{Target|HI} & <  {\theta}_{Target|LI}          \label{eq:t-tt} \\
{\theta}_{Target} & >  \{{\theta}_{Competitor_1}, {\theta}_{Competitor_2}, \theta_{Failure}\}           \label{eq:t-tc}\\
{\theta}_{Competitor_{\{1,2\}}|HI} & >  {\theta}_{Competitor_{\{1,2\}}|LI} \label{eq:t-cc}
\end{align} 

In addition, the direct-access model includes the probability of backtracking,
 $\theta_b$, and the time needed for backtracking, $t_b$.

}

We
provide the estimates of the previous key parameters (or relations between key
parameters) with their credible intervals.

\subsection{Cross-validation} 
We also compared the models using cross-validation, since the descriptive
adequacy can also be achieved by a model that is too flexible and can generate
too many different results. The idea behind cross-validation is to assess the
accuracy the model would have in making predictions for new data, that is the
expected predictive performance.

We compare the  expected predictive performance of the models
\cite{GelmanEtAl2014understanding}
using k-fold cross-validation \cite{VehtariOjanen2012,VehtariEtAl2017}, with k set to
ten. We calculated the k-fold cross-validation by first splitting the data
into ten subsets (or folds) and then using  each subset   as the validation
set, while the remaining data were used for parameter estimation. We
partitioned the data into subsets by pseudo-randomly permuting the
observations, and then systemically dividing them into 10 subgroups; we
ensured that each group contained similar number of observations for each
subject (this was meant to avoid the situation where most of the data of a
certain subject is left out due to chance). The estimate of the expected log
pointwise predictive density ($\widehat{elpd}$) for a new dataset (i.e., the sum
of the expected log pointwise predictive density of each observation)  can be
used as a measure of predictive accuracy for the total number of observation
(N) taken $N/k$ at a time; $\widehat{elpd}$ can be transformed to deviance scale
by multiplying it by minus two, providing a fully Bayesian alternative to AIC
\cite<Akaike Information Criterion;>[]{Akaike1974} or DIC
\cite<Deviance Information Criterion;>[]{SpiegelhalterEtAl2002}.

\subsection{Results}

\subsubsection{Activation-based model}

\begin{figure}[p]
\begin{knitrout}
\definecolor{shadecolor}{rgb}{0.969, 0.969, 0.969}\color{fgcolor}

{\centering \includegraphics[width=.97\linewidth]{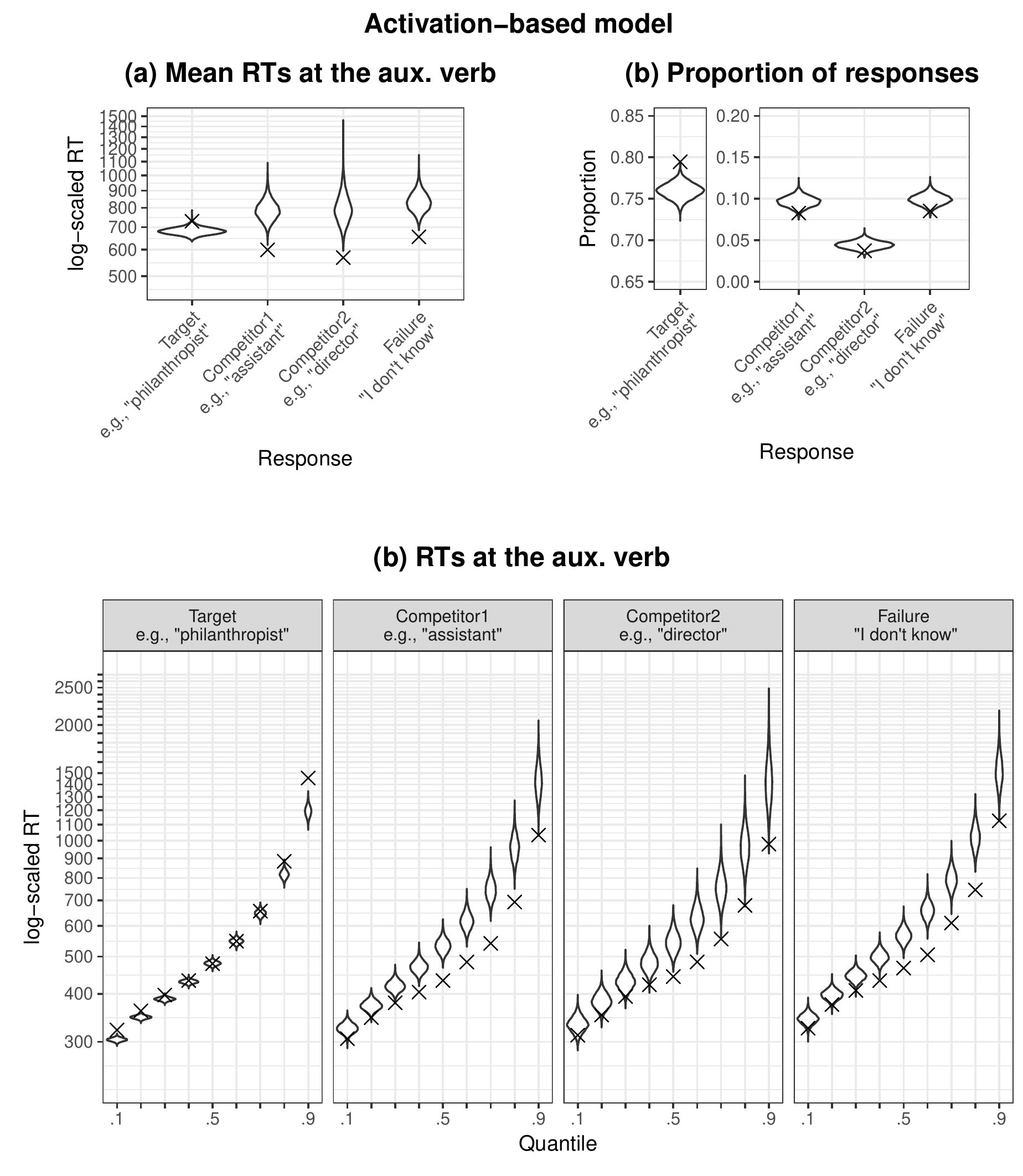} 

}

\end{knitrout}

\caption{The top-most figure shows the fit of the mean reading times (RTs) for
response (a) and proportion of responses (b) of the activation-based model.
The width of the violin plot represents to the density of predicted mean RTs
(a) and responses (b) generated by the model. The bottom figure (c) shows the
fit of the $.1$--$.9$ quantiles of the  reading times (RTs) for response of the
activation-based model.  The width of the violin plot represents to the
density of predicted quantile generated by the model. The observed means and
quantiles are represented with a cross.}\label{fig:generalfit-race-ACTR}
\end{figure}

\paragraph{Posterior predictive check}
The posterior predictive check reveals that the model is inadequate for predicting
some key characteristics of the data. Figure~\ref{fig:generalfit-race-ACTR}(a)
shows that the model predicts shorter times for reading the auxiliary verb
when the correct response is given and longer times for reading the auxiliary
verb when an incorrect answer is given. In other words, the model
underestimates the retrieval time of the correct dependent and overestimates
the retrieval time of the competitor NPs, or the  timeout.
Figure~\ref{fig:generalfit-race-ACTR}(b) also shows  a slight misfit for the predicted
accuracy: the model tends to underestimate the proportion of correct responses
and to slightly overestimate the proportion of incorrect ones. Furthermore,
Figure~\ref{fig:generalfit-race-ACTR}(c) reveals that the fit is especially poor
for the second half of the quantiles.

\paragraph{Estimation of relevant parameters}

\begin{figure}[!htbp]
\begin{knitrout}
\definecolor{shadecolor}{rgb}{0.969, 0.969, 0.969}\color{fgcolor}

{\centering \includegraphics[width=.97\linewidth]{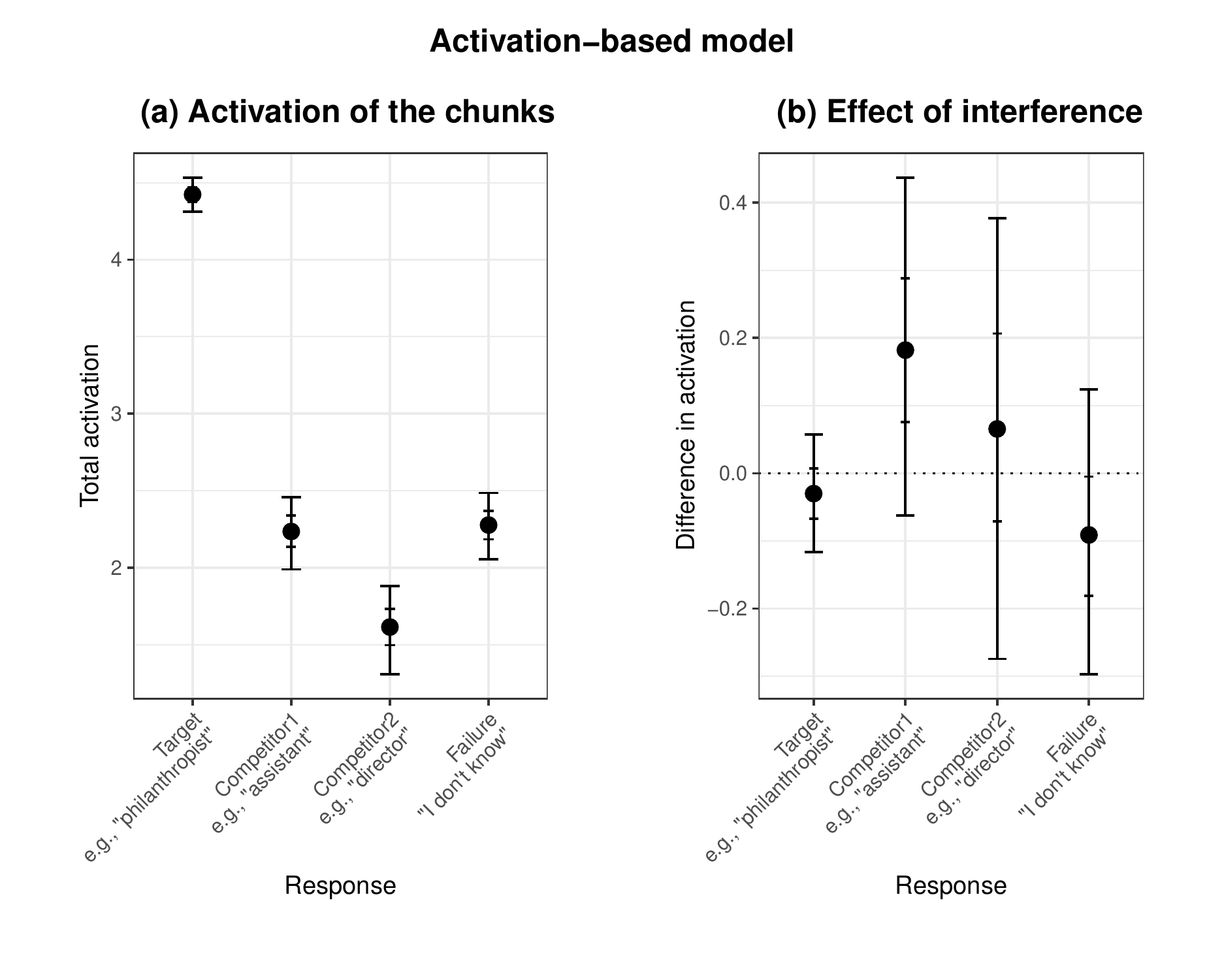} 

}

\end{knitrout}
\caption{Mean activation of the different chunks assuming an arbitrary
threshold of 10 (a), and mean difference between the activations of the chunks
in high interference vs.\ low interference conditions (b). The outer error
bars indicate 95\% credible intervals while the inner error bars indicate 80\%
credible intervals.}\label{fig:pars-race-ACTR}
\end{figure}

The key parameters and relationships between parameters of the
activation-based model are summarized in Figures~\ref{fig:pars-race-ACTR}(a)
and (b). Figure~\ref{fig:pars-race-ACTR}(a) shows caterpillar plots of the
posterior distributions for  the rates of accumulation of evidence for each
choice assuming an arbitrary threshold of 10. In the activation-based model,
these parameters represent the mean activation (together with a common
additive constant) of the target,  competitor NPs, and in the case of the
failure option, the activation represents the speed of the timeout. As assumed
by the activation-based model, the activation of the target is higher than the
activation of the competitors and of the failure. The figure shows that the
activations of the chunks fit the
Equation~(\ref{eq:a-tc}), which indicates that the
correct chunk should receive more activation on average than the competitor
chunks.

\label{p:pars-ab} { {Figure~\ref{fig:pars-race-ACTR}(b)
shows the difference in activation for high and low interference for the target
and the competitors. The figure shows that the posterior means and most of the
probability mass of the effect of interference fit the predictions of
Equations~\eqref{eq:a-tt}, i.e., interference lowers the activation of the
target, and \eqref{eq:a-cc}, i.e., interference raises the activation of the
competitors. However, all the 95\% CrIs include 0 as a plausible value; for the
target: $\hat \beta = -0.03 $, 95\% CrI = $[ -0.12 ,
     0.06 ] , P(\beta>0) =0.25$, for the first competitor: $\hat \beta = 0.18 $, 95\% CrI = $[ -0.06 ,
     0.44 ] , P(\beta>0) =0.93$, and
for the second competitor: $\hat \beta = 0.07 $, 95\% CrI = $[ -0.27 ,
     0.38 ] , P(\beta>0) =0.66$. There is clearly a
lot of uncertainty about the parameters' plausible values. While the posteriors
do not clearly contradict the predictions of the model regarding the effect of
interference, they do not bring much support either. }}

\subsubsection{Direct-access model}

\begin{figure}[p]
\begin{knitrout}
\definecolor{shadecolor}{rgb}{0.969, 0.969, 0.969}\color{fgcolor}

{\centering \includegraphics[width=.97\linewidth]{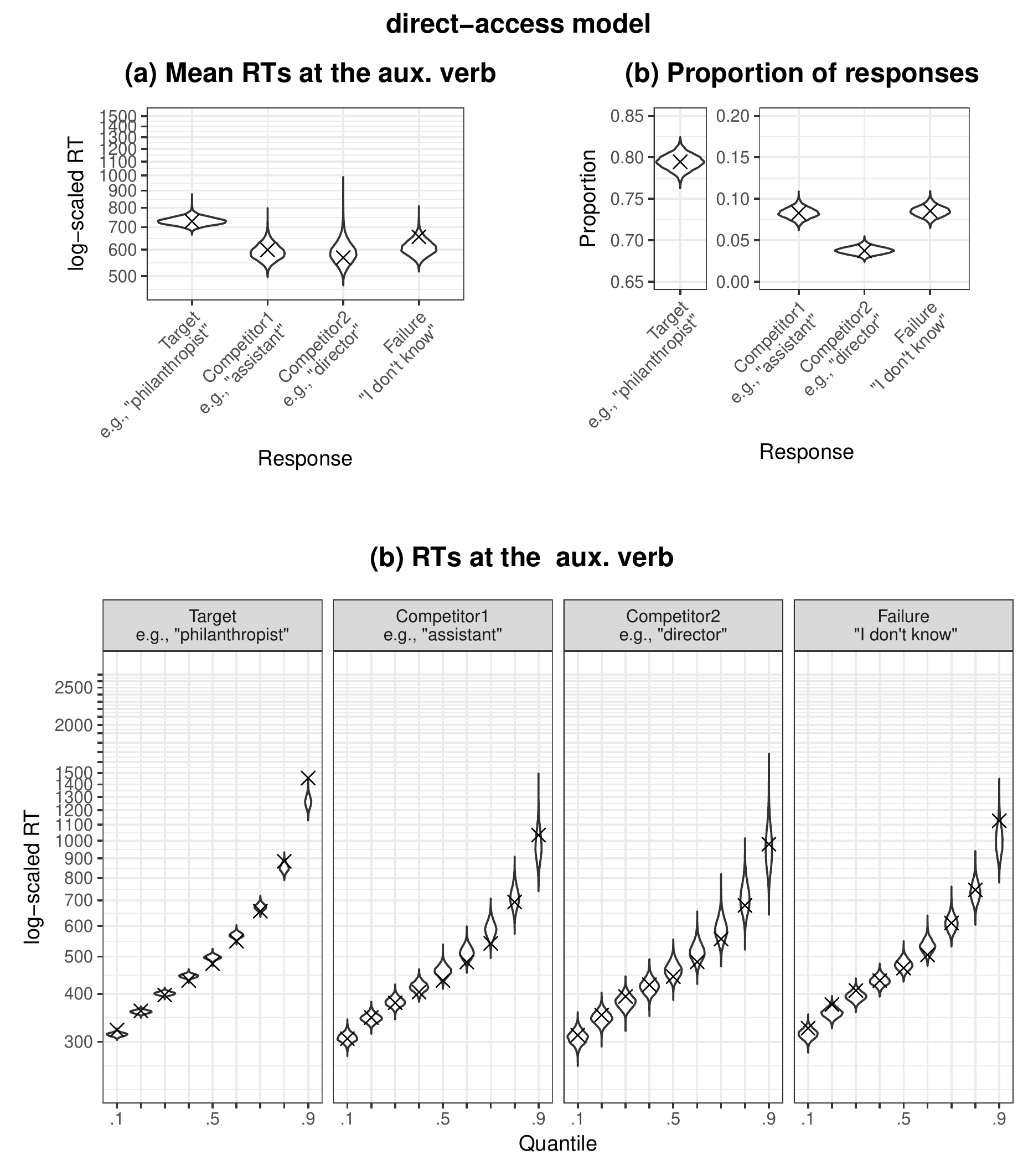} 

}

\end{knitrout}
\caption{The top-most figure shows the fit of the mean reading times (RTs) for
response (a) and proportion of responses (b) of the direct-access model.
The width of the violin plot represents to the density of predicted mean RTs
(a) and responses (b) generated by the model. The bottom figure (c) shows the
fit of the $.1$--$.9$ quantiles of the  reading times (RTs) for response of the
direct-access model.  The width of the violin plot represents to the
density of predicted quantile generated by the model. The observed means and
quantiles are represented with a cross.}\label{fig:generalfit-da-backtracking}
\end{figure}

\paragraph{Posterior predictive check}
The posterior predictive check reveals that, in contrast to the
activation-based model, the direct-access model is able to predict 
the main characteristics of the data fairly well. 
Figure~\ref{fig:generalfit-da-backtracking}(a) shows that the model is able to
predict that reading times associated with correct responses are on average
slower than the ones associated with incorrect ones, while
Figure~\ref{fig:generalfit-da-backtracking}(b) shows that the model is able to
predict fairly well the proportion of responses from the data. Furthermore,
Figure~\ref{fig:generalfit-da-backtracking}(c) reveals that the fit is generally
good for the entire distribution of reading times.

\paragraph{Estimation of relevant parameters}

\begin{figure}[ht!]
\begin{knitrout}
\definecolor{shadecolor}{rgb}{0.969, 0.969, 0.969}\color{fgcolor}

{\centering \includegraphics[width=.97\linewidth]{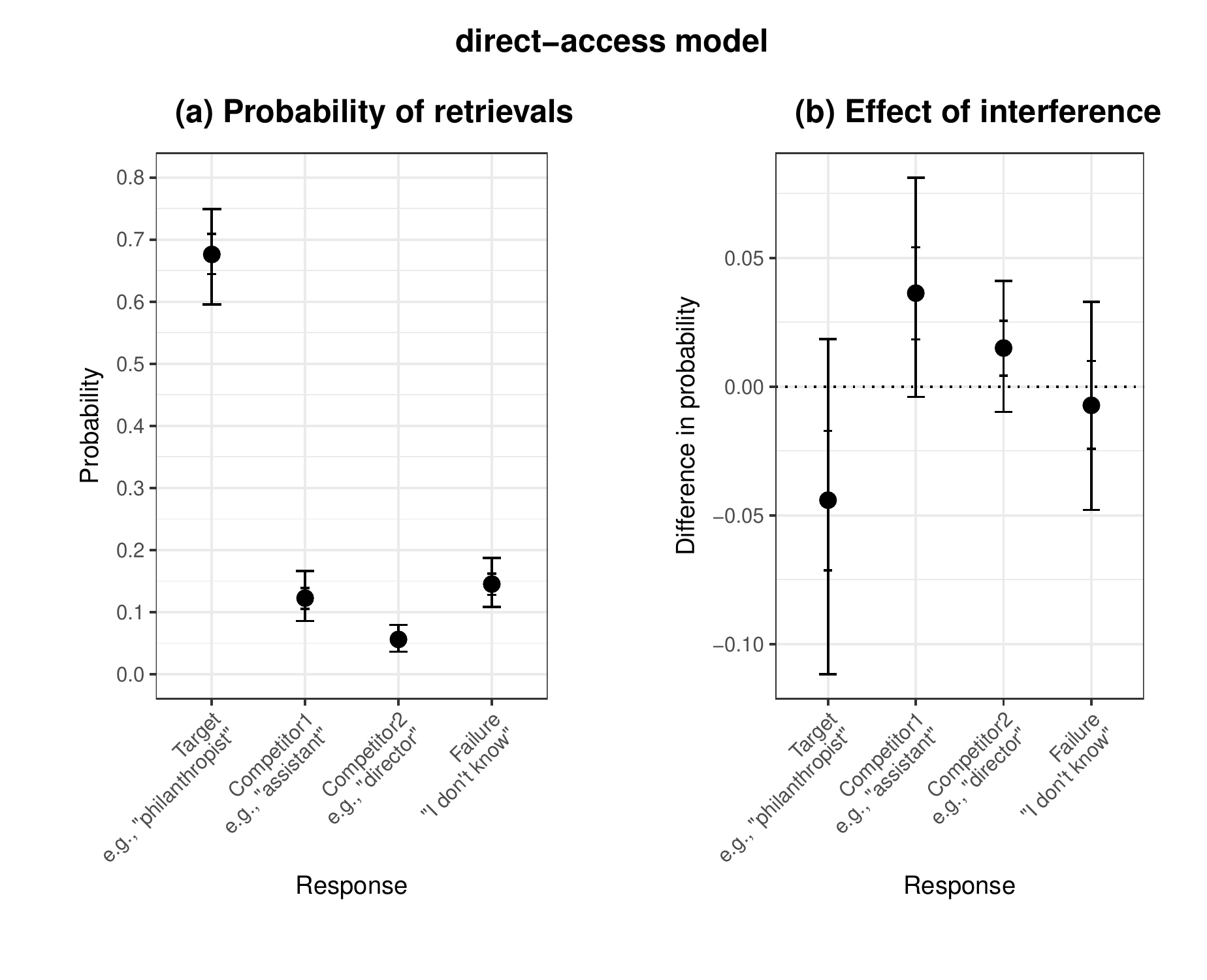} 

}

\end{knitrout}
\caption{Mean probability of retrieval of the different chunks (a), and mean difference between the probabilities due to interference (b). The outer error bars indicate 95\% credible intervals while the inner error bars indicate 80\% credible intervals.}\label{fig:pars-da-backtracking}
\end{figure}

The key parameters of the direct-access model are: (i) the probability that
each of the candidate NPs would be retrieved (as shown in
Figure~\ref{fig:pars-da-backtracking}), (ii) the probability of backtracking
(reported below), and (iii) the time needed for backtracking (reported below).
Figure~\ref{fig:pars-da-backtracking}(a) shows caterpillar plots of the
posterior distributions for the parameters that represent probability of
retrieving each chunk from memory in order to build a dependency at the
auxiliary verb. Figure~\ref{fig:pars-da-backtracking}(a) shows that the
retrieval of the target is more likely than the retrieval of the competitors
or the retrieval failure; this is in agreement with Equation
\eqref{eq:t-tc}. Notice that since the model assumes that
backtracking is possible, after some trials the incorrect retrieval will be
repaired. This means that the probability of retrieving a dependent is not the
same as the proportion of times  a response was given in  the multiple choice
task.  In fact, the model estimates that around half of the time that there is
a misretrieval, it will be corrected ($\hat \theta_b = 0.48 $, 95\% CrI = $[ 0.4 ,
     0.55 ] $). In addition, the
model estimates that backtracking takes
120 ms, $95\%$ CrI $=[30,
231 ]$ ms (after transforming it from log-scale).

{ {As shown in 
Figure~\ref{fig:pars-da-backtracking}(b) and similarly to the case of the
activation-based model, the posterior distributions for the effect of
interference do not clearly support or contradict the predictions of the model
regarding the effect of interference. As before, the means  and most of the
probability mass of the posteriors for the effect of interference on the target
and competitors fit the predictions of Equations~\eqref{eq:t-tt}, i.e.,
interference lowers the probability of retrieval of the target,  and
\eqref{eq:t-cc}, i.e., interference raises the probability of retrieval of the
competitors. However, all the 95\% CrIs include 0; for the target:
$\hat \beta = -0.04 $, 95\% CrI = $[ -0.11 ,
     0.02 ] , P(\beta>0) =0.08$, for the first competitor: $\hat \beta = 0.04 $, 95\% CrI = $[ 0 ,
     0.08 ] , P(\beta>0) =0.96$, and for the
second competitor: $\hat \beta = 0.02 $, 95\% CrI = $[ -0.01 ,
     0.04 ] , P(\beta>0) =0.88$. }}\label{p:pars-da}

\subsubsection{Cross-validation: activation-based vs.\ direct-access models}

In order to assess the compatibility of the models with the data, we compared
how  the models would generalize to an independent data set, that is, the
pointwise out-of-sample prediction accuracy or $\widehat{elpd}$ of the models.
Comparing the models using k-fold cross validation reveals an estimated
difference in $\widehat{elpd}$ of $-110$
($SE=28$) in favor of the direct-access model
($\widehat{elpd}= -26747$,
$SE=100$) in comparison with the
activation-based model ($\widehat{elpd}=
-26858$,
$SE=100$).

\begin{figure}[ht!]
\begin{knitrout}
\definecolor{shadecolor}{rgb}{0.969, 0.969, 0.969}\color{fgcolor}

{\centering \includegraphics[width=.5\linewidth]{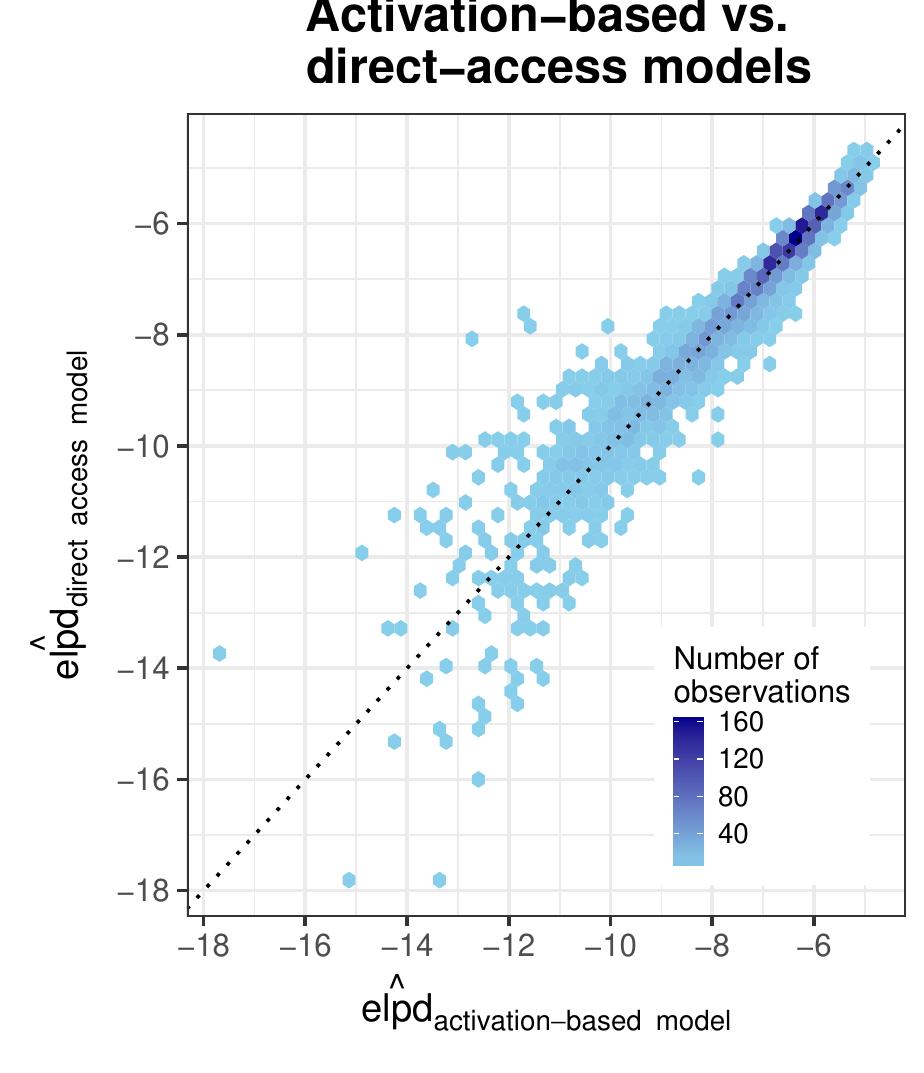} 

}

\end{knitrout}
\caption{Comparison of the activation-based and direct-access models in terms
of their predictive accuracy for each observation. Each axis shows the
expected pointwise contributions to 10-fold cross-validation for each model
($\widehat{elpd}$ stands for the expected log pointwise predictive density of
each observation). Higher (or less negative) values  of $\widehat{elpd}$ indicate
a better fit. Darker cells represent a higher concentration of observations
with a given fit.}\label{fig:fit-CV-1}
\end{figure}

\begin{figure}[ht!]
\begin{knitrout}
\definecolor{shadecolor}{rgb}{0.969, 0.969, 0.969}\color{fgcolor}

{\centering \includegraphics[width=.97\linewidth]{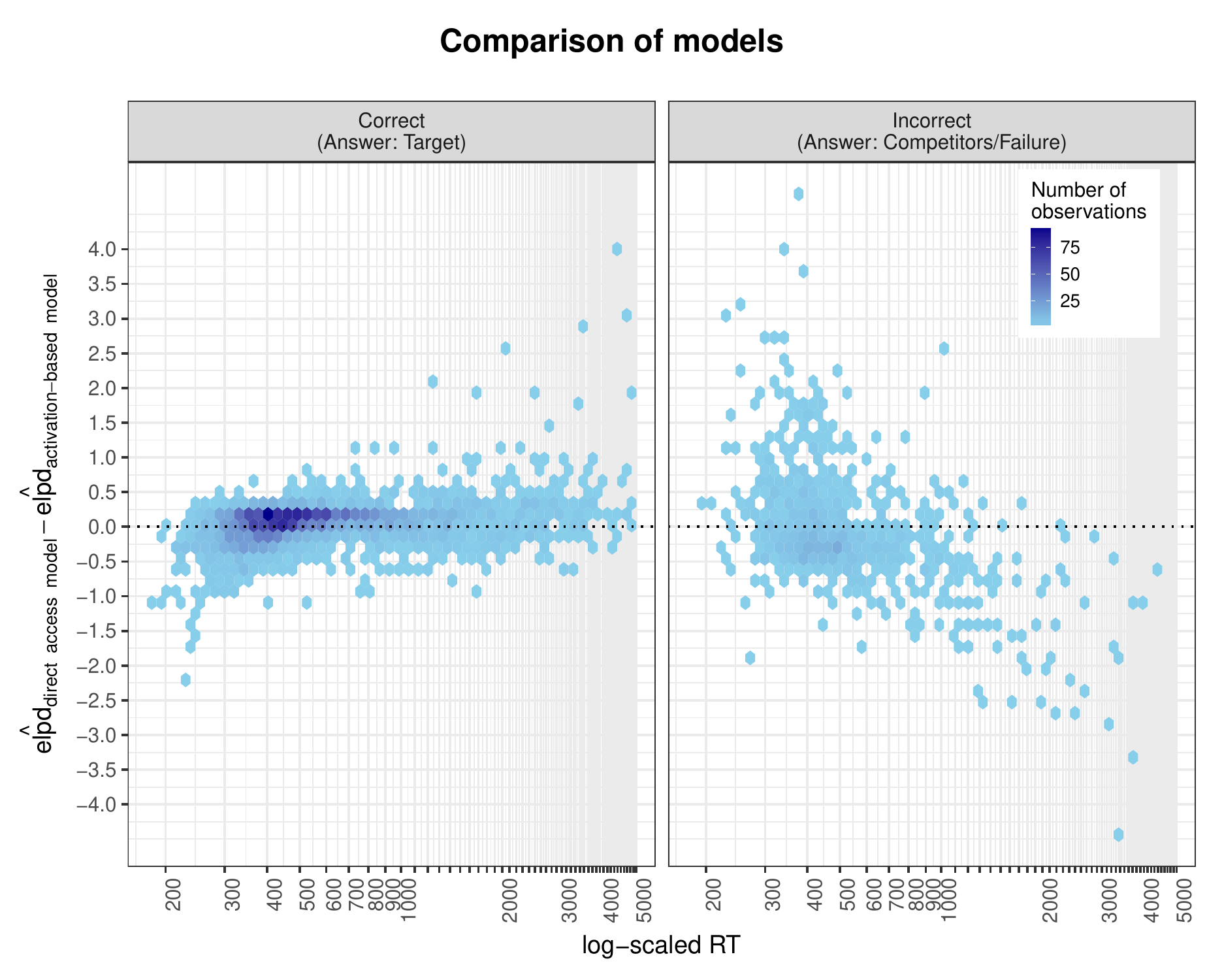} 

}

\end{knitrout}
\caption{Comparison of the activation-based and direct-access models in terms
of their predictive accuracy for each observation depending on its
log-transformed reading time (x-axis) and accuracy (left panel showing correct
responses, and the right panel  showing  any of the possible incorrect
responses). The y-axis shows the difference between the expected pointwise
contributions to 10-fold cross-validation for each model ($\widehat{elpd}$ stands
for the expected log pointwise predictive density of each observation); that
is,  positive values represent  an advantage for the direct-access model while
negative values represent an advantage for the activation-based model. Darker
cells represent a higher concentration of observations with a given
fit.}\label{fig:fit-CV-2}
\end{figure}

Figure \ref{fig:fit-CV-1} shows for any given observation, whether one model
has an advantage over the other in its predictive accuracy. Since higher (or
less negative) values of $\widehat{elpd}$ indicate a better fit for a model,
observations that are further away from the dotted line correspond to data
that are particularly better predicted by one model (and poorly by the other).
This figure shows that the advantage of the direct-access model is not due to
some outlier observations, but mostly due to a high number of observations
that  fit slightly better under this model than under the activation-based one
(this is the darker patch on the top right corner). Figure
\ref{fig:fit-CV-2} shows the difference between the $\widehat{elpd}$ of the two
models for every observation corresponding to either a correct or an incorrect
response. The figure shows that most of the advantage of the direct-access model comes from reading times between 300 and 1000 ms (notice the darker
patch above the zero dotted line). In addition, the direct-access model has a clear
advantage in predicting  long reading times associated with correct responses
and short reading times associated with incorrect ones, while  the
activation-based model has an advantage in predicting short reading times for
correct responses and long reading times for incorrect ones.

\newpage

\subsection{Discussion}

The evaluation of the activation-based and direct-access models reveals two
sets of findings: one relates to the effect of interference on
the key parameters of the models, and other the relates to their
validity as models of retrieval in sentence comprehension.

Regarding the effect of interference on the parameters of the models, the
results show that  interference affects the parameters as expected, but some
of the posteriors include a large degree of uncertainty. Given the relatively small
effect of interference in reading times in the experimental study of
\citeA{NicenboimEtAl2016NIG}, and given that we used a subset of the original
data, this is not surprising. However, this serves as a sanity check that
confirms that both models can work in principle and that experimental findings
can produce the expected effects on the parameters of the models. 

For the
activation-based model, the underlying activation of the target NP was, as
expected, clearly larger on average than the one of the competitors and the
one associated with the timeout. The parameters that correspond to the effect
of interference on activation, however, provided very weak evidence that
interference decreases the activation of the target and increases the
activation of  the competitors. 

Similarly for the direct-access model, the underlying probability of retrieving
the target was clearly larger than the one of the competitors and the one
associated with the failure of the retrieval process. The model estimated that
approximately half of the time ($\hat \theta_b = 0.48 $, 95\% CrI = $[ 0.4 ,
     0.55 ] $) that a
retrieval was incorrect, it was repaired to the correct retrieval in
120 ms, $95\%$ CrI
$=[30, 231 ]$
ms. This finding shows, as \citeA{McElree2000} suggested,  that it is
possible to account for differences in reading times that arise only from
differences in probabilities of retrieval, if there is a repair process that
takes more than a negligible amount of time. However, as with the
activation-based model, the posterior distributions present only weak evidence
that interference decreases the probability of retrieving the target and
increases the probability of retrieving one of the competitors.

In order to evaluate the validity of the models for retrieval in sentence
comprehension, we examined whether the models were able to fit the patterns
found in the data using posterior predictive checks, and we compared their
predictive accuracy using cross-validation. The posterior predictive checks of
the activation-based and direct-access models show clearly that some aspects of
the data fit better under the direct-access model than under the
activation-based model. While the reading times at the auxiliary verb associated
with correct responses in the multiple choice task were on average slower than
the reading times associated with incorrect responses, this pattern could only
be captured by the direct-access model. This is so because in the case of the
direct-access model, reading times associated with correct responses are assumed
to be a mixture of fast direct accessed retrievals and slower backtracked
responses, while incorrect responses are assumed to be just direct accessed
wrong or failed retrievals. By contrast, in the case of the activation-based
model, reading times associated with correct responses are assumed to be faster
on average than reading times associated with incorrect responses. The
activation-based model assumes a race between the accumulation of evidence for
the candidates to the retrieval, where the fastest item is the one retrieved.
The particular characteristics of this race-between-accumulators model, which
are motivated by ACT-R, include the assumption of a ballistic race
\cite<lack of fluctuations occurring during the accumulation process or
within-choice noise;>[]{BrownHeathcote2005ballistic}, the same variance
parameter for all the accumulators (i.e., a single between-choice noise),
{ and no variation in the initial bias
\cite{HeathcoteLove2012}}. Under this type of race,  the correct responses,
which are answered more frequently than the incorrect ones, will also be the
fastest on average.

Even though  reading times for correct responses were on average slower than
the ones for incorrect ones, this was not the case for every observation.
Model comparison using cross-validation shows that the advantage of the
direct-access model is based mainly on giving a better fit for reading times
between 300 and 1000 ms, while the model is worst suited to predict fast
reading times corresponding to correct responses and slow reading times
corresponding to incorrect responses, which are better predicted by the
activation-based model (see Figure \ref{fig:fit-CV-2}).

Cross-validation supports the direct-access model, but does not
rule out  others models that assume a race between accumulators of evidence
for each retrieval candidate: the concept of activation determining retrieval
latencies and accuracy may still be fruitful. It may be possible to explain
the pattern in the data by including a mixture process in the race model, that
is, if it is assumed, in a similar way as with the direct-access model,  that
the reading times associated with the correct responses are a mixture of fast
retrievals due to high activation together with repaired wrong or failed
retrievals. However, a model like this would be  too flexible for the data at
hand and may present problems of identifiability (since it would be hard to
estimate the activation of the non-retrieved candidates).

A closely related model that has been proposed to account for fast errors  by
\citeA{NicenboimEtAl2016Frontiersb} assumes that failed retrievals may take
less time than completed retrieval. This is achieved by assuming that, when the
activation is too low, the retrieval is aborted instead of waiting until the
timeout is reached. However, this would  only explain the fast failures (``I
don't know answers'' in the multiple choice task), but it would still leave
fast retrievals of competitor NPs unexplained.

As we mentioned before, the activation-based model is based on a specific race
model, namely the lognormal race model, which in turns is a very specific
implementation of a model that assumes the sequential sampling of evidence for
a decision \cite<a class of models that includes the race of accumulators
and random walk/diffusion models; for a review, see>[]{RatcliffEtAl2016}. There
are other tasks that trigger error responses that are on average faster
than correct responses and have been explained with sequential
sampling models such as the drift diffusion model
\cite{RatcliffRouder1998,WagenmakersEtAl2008}, or the linear ballistic
 accumulator \cite{BrownHeathcote2008}.\label{p:fast} {
{ These models, in contrast with the lognormal race model, allow
for variation in the initial bias due to noise. Fast errors can be captured by
assuming a low threshold of evidence for all the decisions (in comparison with
the variation in the initial bias). The low thresholds will produce faster
responses in general, but the increase in speed will be larger for error
responses. This is because incorrect responses will to tend occur when the process
starts near the error boundary for the drift diffusion model or near the
thresholds of evidence of the incorrect responses for the linear ballistic
accumulator \cite{WagenmakersEtAl2008,HeathcoteLove2012}. This is in
contrast with the situation where the thresholds of evidence are higher in
comparison with the initial bias, in this case errors will tend to occur  when
the target candidate was slower than average.  Since models such as the drift
diffusion model or the linear ballistic accumulator allow fast or slow errors
depending on the participants' decision thresholds, an interesting implication
of the fast errors in the current experiment would be that participants
privileged speed at the expense of accuracy without explicit instructions.
}}Even though  the aforementioned models could account for the faster average
reading times associated with incorrect responses,  they would lose the close
connection with the ACT-R framework that motivated our use of the lognormal race
model.

\begin{figure}[p]
\begin{knitrout}
\definecolor{shadecolor}{rgb}{0.969, 0.969, 0.969}\color{fgcolor}

{\centering \includegraphics[width=.97\linewidth]{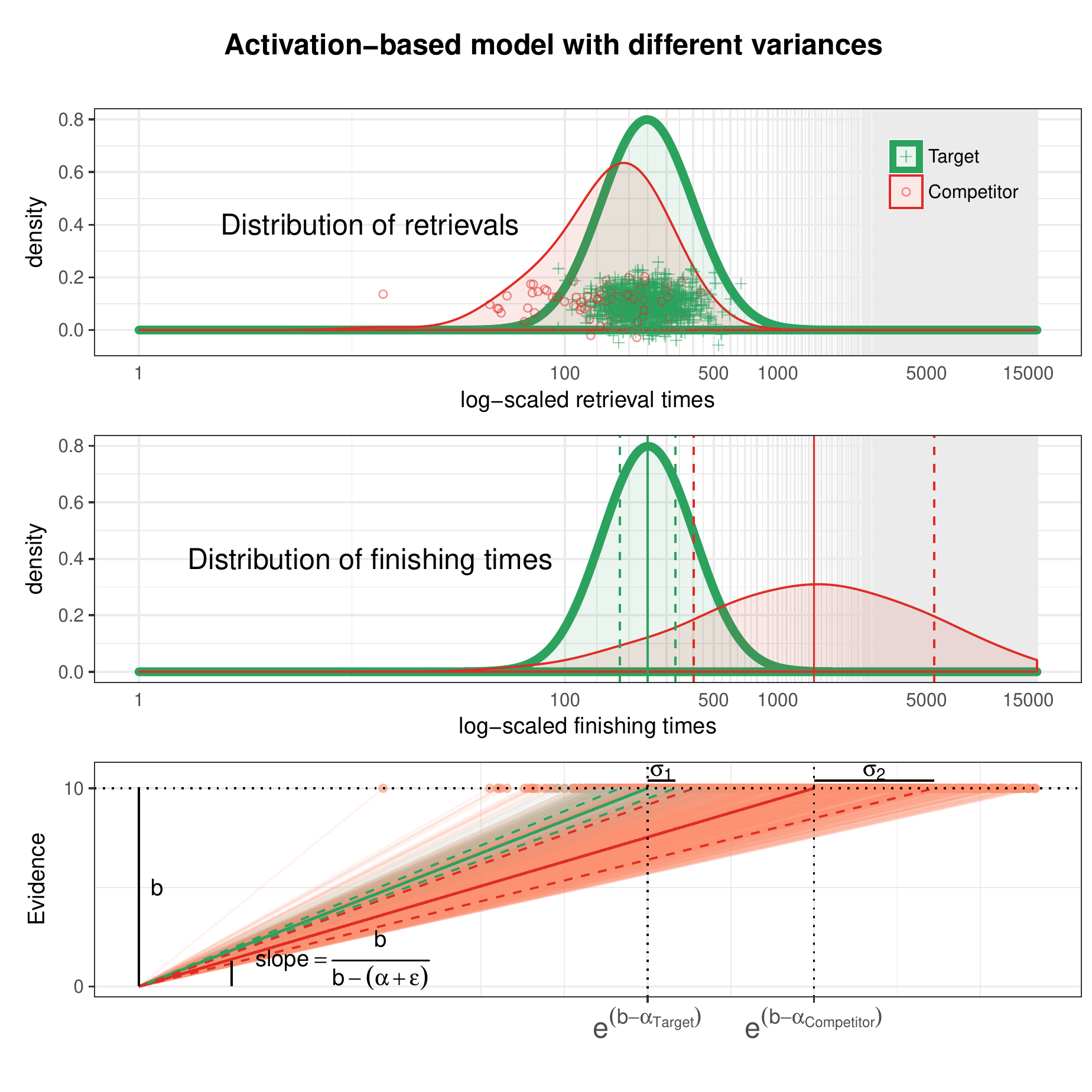} 

}

\end{knitrout}
\caption{The figure depicts how the distribution of retrievals is generated
from the activation-based model with different variances. The bottom figure
depicts the parameters of the activation-based model, the full lines in green
and red are the mean finishing times $t_{c=Target}$ and $t_{c=Competitor}$,
and the broken lines are finishing times one standard deviation away from the
mean. The middle figure shows the distributions of finishing times for target
and candidate; since every chunk is associated a potential finishing time,
$t$, both distributions have the same number of elements. The top-most figure
shows the distribution of retrieval times (adding the shift parameter $\psi$
would transform it to reading times); since only the winning chunks are
retrieved, the distribution of retrieval times for targets has more elements
than the distribution of the competitors.  Notice that even though the finishing times
for the competitors are slower on average than those of the targets (middle
plot), the situation is reversed for the retrieval times (top-most plot).
}\label{fig:race-multnoise}
\end{figure}

Regarding the lognormal race model, its limitation is that if equal variance
is associated with each accumulator, fast errors on average cannot be
predicted because bias (distance) and rate of
accumulation cannot be disentangled
\cite{HeathcoteLove2012,RouderEtAl2014}. Fast errors on average can be
predicted, if the variance of the accumulators of the incorrect
responses is larger than the one of the correct response. 
\citeA{HeathcoteLove2012} propose that poorer matches may spread not only
weaker activation on average but may also be noisier than stronger matches.
This idea is also present in SAM framework of \citeA{GillundShiffrin1984},
which assumes slower reactions and more variability for  poorer matches than
for more precise matches. Figure \ref{fig:race-multnoise} shows graphically
how the distribution of correct and incorrect retrievals is generated from the
activation-based model with different variances.

In the following section, we evaluate the activation-based model with
different variances, one for the accumulator of target and one for the other
accumulators, and we compare it with the direct-access model.

\section{Evaluation of the activation-based model with different variances}

We evaluated the activation-based model with different
variances using the same data as with the previous models. As before, we (i)
examined the descriptive adequacy of the model using posterior predictive
checks, (ii) estimated its relevant parameters, and (iii) compared it with
the direct-access model using cross-validation.

The assumptions of the activation-based model with different variances are
identical to those of the default activation-based model, except that the
noise in the rate of accumulation of evidence of each chunk can have different
variances. This means that the lognormal distributions associated with each
activation have different scale parameters (which corresponds to the standard
deviation of the associated normal distribution). Since all the competitors
were retrieved only 21\% of the time, for simplicity (and for
improving the convergence of the models) we assumed only two variances, one for
the lognormal distribution associated with the target, and one for the
competitors or the failure timeout.

\subsection{Results}

\begin{figure}[p]
\begin{knitrout}
\definecolor{shadecolor}{rgb}{0.969, 0.969, 0.969}\color{fgcolor}

{\centering \includegraphics[width=.97\linewidth]{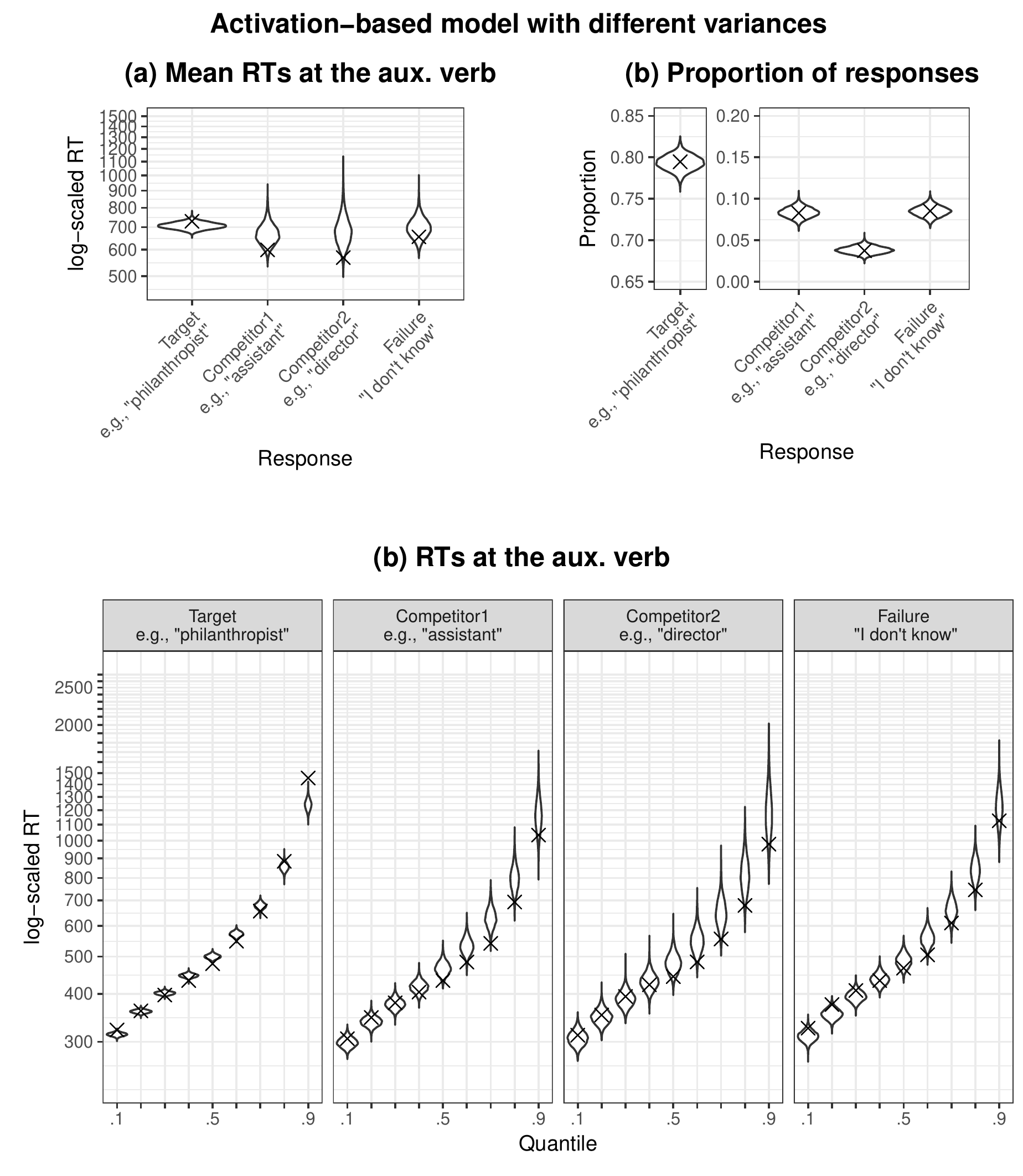} 

}

\end{knitrout}
\caption{
The top-most figure shows the fit of the mean reading times (RTs) for
response (a) and proportion of responses (b) of the activation-based model with different variances.
The width of the violin plot represents to the density of predicted mean RTs
(a) and responses (b) generated by the model. The bottom figure (c) shows the
fit of the .1-.9 quantiles of the  reading times (RTs) for response of the
activation-based model with different variances.  The width of the violin plot represents to the
density of predicted quantile generated by the model. The observed means and
quantiles are represented with a cross.}\label{fig:generalfit-noisy-activation}
\end{figure}

\paragraph{Posterior predictive check}
The posterior predictive check reveals that the activation-based model with different variances can capture the main characteristics of the
data. Figure~\ref{fig:generalfit-noisy-activation}(a) shows that the
model predicts a wide range of reading times associated with the incorrect
responses, and  most of the predicted reading times associated with incorrect
responses are only slightly faster than the correct ones.
Figure~\ref{fig:generalfit-noisy-activation}(b) shows that the model is able
to predict the proportion of responses from the data.
Figure~\ref{fig:generalfit-noisy-activation}(c) reveals that the fit is better for the
first half of the quantiles, while for the second half of the quantiles the
predicted data contains the observed quantiles, mainly because of the wide
distribution of predicted reading times.

\paragraph{Estimation of relevant parameters}

\begin{figure}[ht!]
\begin{knitrout}
\definecolor{shadecolor}{rgb}{0.969, 0.969, 0.969}\color{fgcolor}

{\centering \includegraphics[width=.97\linewidth]{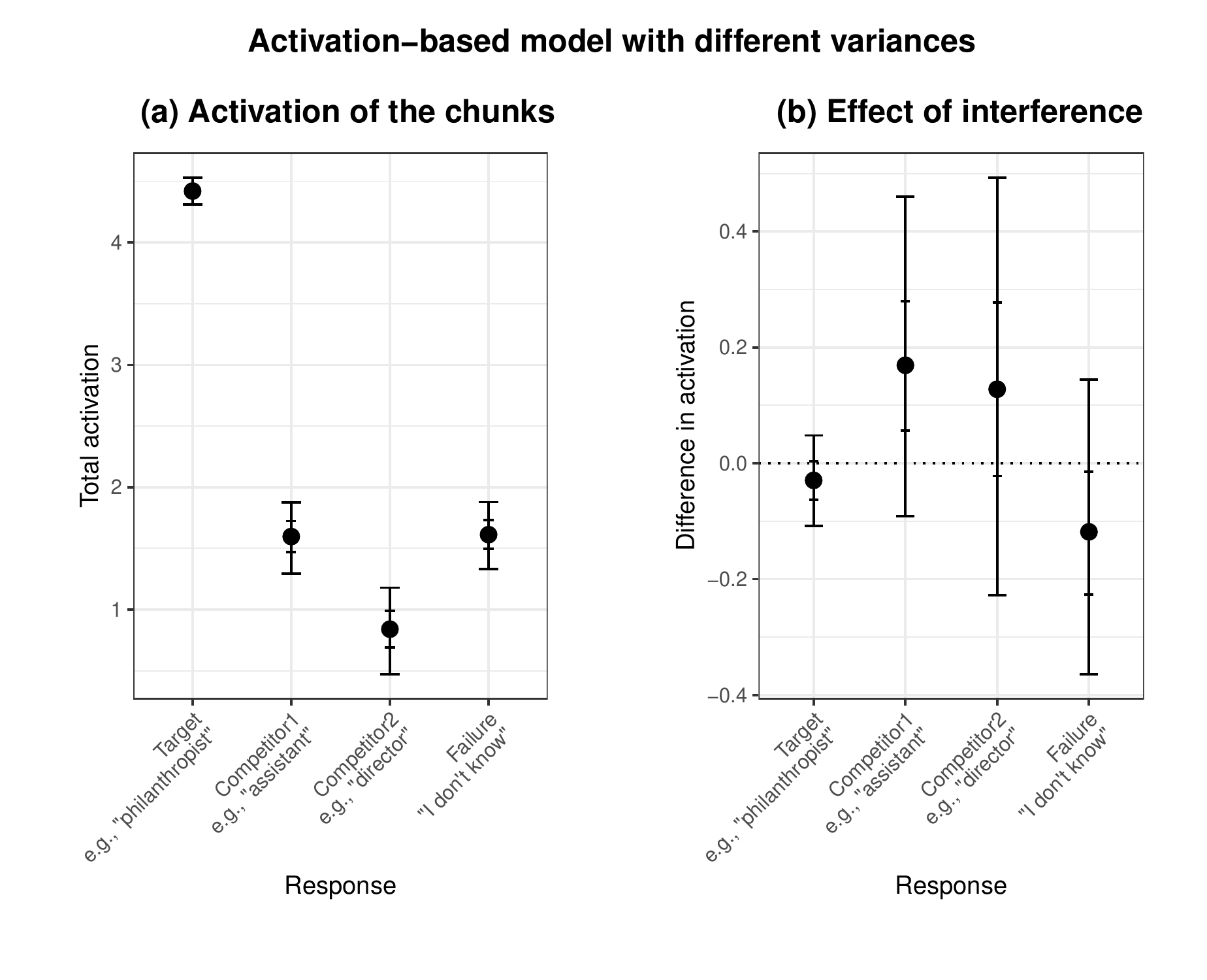} 

}

\end{knitrout}
\caption{Mean activation of the different chunks assuming an arbitrary
threshold of 10 (a), and mean difference between the activations due to
interference (b). The outer error bars indicate 95\% credible intervals while
the inner error bars indicate 80\% credible
intervals.}\label{fig:pars-noisy-activation}
\end{figure}

\begin{figure} 
\begin{knitrout}
\definecolor{shadecolor}{rgb}{0.969, 0.969, 0.969}\color{fgcolor}

{\centering \includegraphics[width=.5\linewidth]{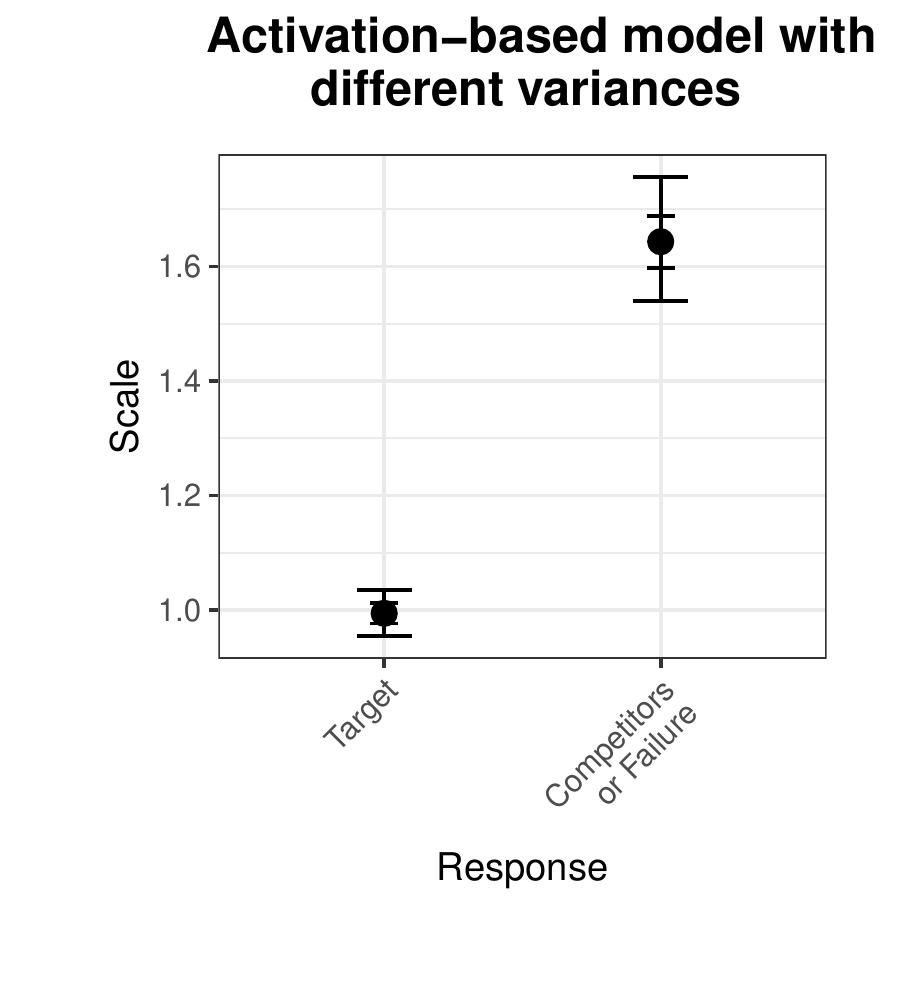} 

}

\end{knitrout}
\caption{The figure depicts that the scale of the distributions of activations
of the target chunk and of the competitors or timeout. The outer error bars
indicate 95\% credible intervals while the inner error bars indicate 80\%
credible intervals.}\label{fig:sigmas-noisy-activation}
\end{figure}

The estimation of the key parameters and the relationship between parameters of
the activation-based model with different variances shows similar results to the
ones in the default activation-based model. As in the default model,
Figure~\ref{fig:pars-noisy-activation}(a) shows that the activation of the
target is higher than the activation of the competitors and of the failure. In
addition and also similarly to the case of the default model,
Figure~\ref{fig:pars-noisy-activation}(b) shows that { 
\label{p:pars-ab2}{while the posterior means and most of the
 probability mass of the effect of interference on the target and competitors
 fit the predictions of Equations~\eqref{eq:a-tt} and \eqref{eq:a-cc}, all the
 95\% CrIs include 0; for the target: $\hat \beta = -0.03 $, 95\% CrI = $[ -0.11 ,
     0.05 ] , P(\beta>0) =0.23$, for the first
 competitor: $\hat \beta = 0.17 $, 95\% CrI = $[ -0.09 ,
     0.46 ] , P(\beta>0) =0.89$, and for the second competitor:
$\hat \beta = 0.13 $, 95\% CrI = $[ -0.23 ,
     0.49 ] , P(\beta>0) =0.77$.}} Recall that  the variance was allowed
to be different for the correct and incorrect retrievals;
Figure~\ref{fig:sigmas-noisy-activation} shows that, as hypothesized, this
allows the scale associated with the distribution of activations of the incorrect
retrievals to be  larger than the one associated with correct retrievals.

\subsubsection{Cross-validation: activation-based model with different variances vs.\ direct-access model}

A comparison of the activation-based model with different variances and direct-access model using 10-fold cross-validation shows that the estimates of
$\widehat{elpd}$ are very similar, with a very small advantage for the
activation-based model with different variances ($\widehat{elpd}=
-26725$,
$SE=98$) in
comparison with the direct-access model ($\widehat{elpd}=
-26747$,
$SE=100$), namely an estimated
difference in $\widehat{elpd}$ of $-22$
($SE=18$); while the advantage of the direct-access model in
comparison with the default activation-based model was of
-110 with $SE=28$. { In addition, the
activation-based model with different variances shows a clear advantage in
comparison with the default activation model, namely an estimated difference in
$\widehat{elpd}$ of
$-133$ ($SE=18$)}

\begin{figure} 
\begin{knitrout}
\definecolor{shadecolor}{rgb}{0.969, 0.969, 0.969}\color{fgcolor}

{\centering \includegraphics[width=.5\linewidth]{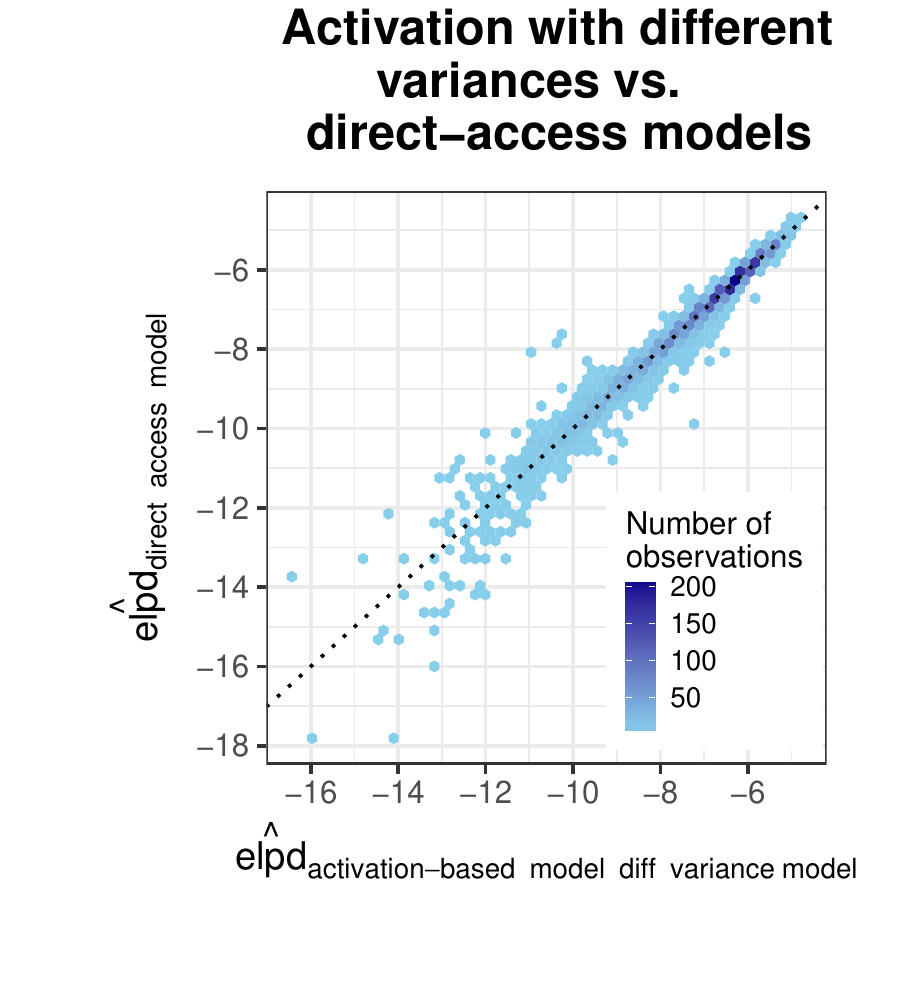} 

}

\end{knitrout}
\caption{Comparison of the activation-based model with different variances and
the direct-access model in terms of their predictive accuracy for each
observation. Each axis shows the expected pointwise contributions to 10-fold
cross validation for each model ($\widehat{elpd}$ stands for the expected log
pointwise predictive density of each observation). Higher (or less negative)
values  of $\widehat{elpd}$ indicate a better fit. Darker cells represent a
higher concentration of observations with a given fit. }\label{fig:fit-CV-3}
\end{figure}

\begin{figure}[ht!]
\begin{knitrout}
\definecolor{shadecolor}{rgb}{0.969, 0.969, 0.969}\color{fgcolor}

{\centering \includegraphics[width=.97\linewidth]{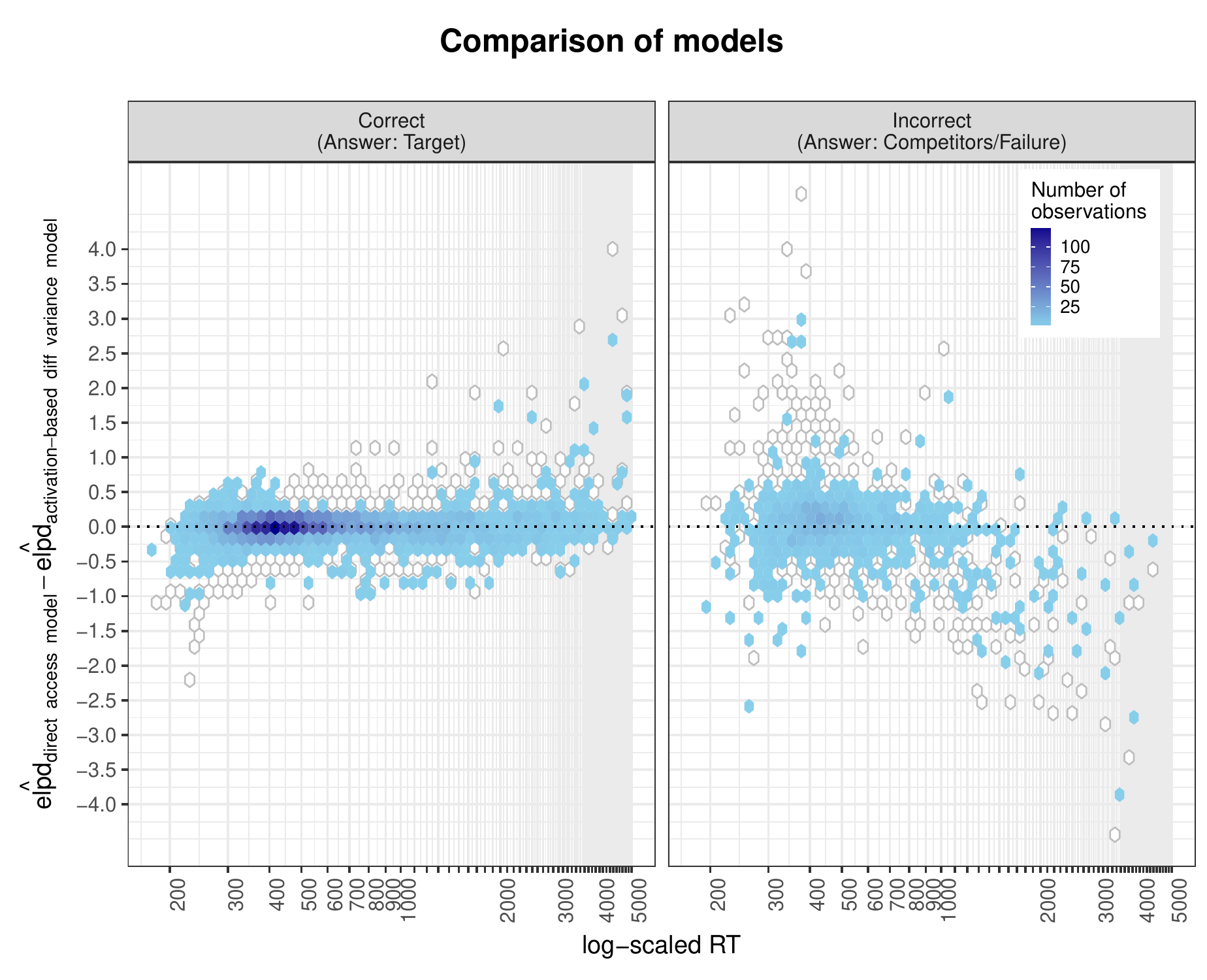} 

}

\end{knitrout}
\caption{Comparison of the activation-based model with different variances
and direct-access model in terms of their predictive accuracy for each
observation depending on its log-transformed reading time (x-axis) or accuracy
(left panel showing correct responses, and the second panel  showing  any of
the possible incorrect responses). The y-axis shows the difference between the
expected pointwise contributions to 10-fold cross-validation for each model
($\widehat{elpd}$ stands for the expected log pointwise predictive density of
each observation); that is,  positive values represent  an advantage for the
direct-access model while negative values represent an advantage for the
activation  model with different variances. Darker cells represent a higher
concentration of observations with a given fit. The white cells show the comparison (shown earlier in Figure
\ref{fig:fit-CV-2}) of the
default activation-based model with the direct-access model.}\label{fig:fit-CV-4}
\end{figure}

 Figure~\ref{fig:fit-CV-3} shows that the predictive accuracy of the models is
fairly similar with most of the observations being fit well by both of them.
There are, however, some observations scattered at the bottom left corner of
Figure~\ref{fig:fit-CV-3}, which favors the activation-based model with
different variances. Figure~\ref{fig:fit-CV-4} shows in blue cells the
difference between the $\widehat{elpd}$ of both models for every observation
corresponding to either a correct or an incorrect response; and in white cells
the previous comparison (from Figure~\ref{fig:fit-CV-2}) of the default
activation-based model and direct-access model. Figure \ref{fig:fit-CV-4} shows
that the difference between the activation-based model with different variances
and direct-access model is smaller than the difference between the direct-access
model and the (default) single-variance activation-based model. The main
difference between the fits is that the activation-based model with different
variances is able to account better for some fast and slow reading times
associated with incorrect responses.

\newpage

\subsection{Discussion}

The estimation of the relevant parameters of the activation-based model with
different variances shows that the scale parameter associated with the
distribution of activations for incorrect retrievals is larger than the one
associated with correct ones, as it is necessary to account for fast errors.
However, this did not change the predicted interference effect compared to 
the default activation-based model. Similarly, as with the default model, the parameters
that correspond to the effect of interference on activation 
{ do not clearly support or 
contradict the predictions of the model regarding the effect of interference, that is 
that interference decreases the activation of the target and
increases the activation of the competitors.}

Regarding the descriptive adequacy of the model, even though the inclusion of
different variances improves the fit, the posterior predictive checks show
more variation on the predicted reading times associated with incorrect
responses for this model than for the direct-access model. This is not
necessarily a disadvantage, and it may indicate that the direct-access model
is more flexible and may be slightly overfitting the data, since these
predictions are generated with the best estimates (and posterior
distributions) to account for the data. In fact, despite an apparent better
fit for the direct-access model, the estimates of predictive accuracy
($\widehat{elpd}$) are very similar with a very slight advantage for the
activation-based model with different variances. In contrast with the
difference between the fit of the original models (i.e., the default
activation-based vs.\ the direct-access model shown in
Figure~\ref{fig:fit-CV-2}), the difference between the fit of the
activation-based model with different variances and the direct-access model is
smaller (see Figure \ref{fig:fit-CV-4}), with the new version of the
activation-based model giving a better fit to some of the fast and slow
reading times associated with incorrect responses.

This comparison shows that even though the inclusion of
different variances for the accumulators does not  imply a clear superiority
over the direct-access model, it is possible to account for the data with a
model which is based on a race of accumulation of evidence.

\section{General discussion}

We evaluated two models that have been successful in explaining similarity-based
interference in sentence comprehension:
\posstextcite{LewisVasishth2005} activation-based model following  ACT-R
assumptions \cite{AndersonEtAl2004}  and
\posstextcite{McElree2000} direct-access model. We also
evaluated a third model, a variation of the activation-based model.

The models were implemented in a Bayesian hierarchical
framework and fit to the data of \citeA{NicenboimEtAl2016NIG}. Even
though the activation-based model was already implemented computationally
\cite{LewisVasishth2005}, our implementation enabled us to go beyond
simulations as they are usually done for this model
\cite<e.g.,>[]{VasishthLewis2006,NicenboimEtAl2016Frontiersb}, and to fit
the observations of an experiment by accounting for variation coming from
participants and experimental items. For the direct-access model, we provided a
first computational implementation  which  allowed us to derive precise and
unambiguous predictions, which are fully transparent in our instantiation of
the model. We first summarize our findings, and we then discuss the motivation
of this work, the implications of the findings, and future work.

Our evaluation can be summarized in three main results. First, the underlying
parameters of both models behave as expected under interference effects.
However, the parameters showed a large degree of uncertainty in their
posterior distributions. While this may be due to the relatively small magnitude of the
interference effect in the original experiment
\cite{NicenboimEtAl2016NIG}, the findings confirm that, as expected,
both models can in principle explain interference effects.

Second, we evaluated the validity of both models in predicting the reading
times and accuracy patterns during retrieval. The posterior predictive checks
and the comparison using cross-validation show  that some aspects of the data
fit better under the direct-access model than under the default
activation-based model. The data showed on average slower reading times
associated with correct responses than with incorrect ones, and this pattern
could be explained only by the direct-access model. This suggests that the
default activation-based model may not be flexible enough to accommodate
patterns in the data that go beyond means between conditions.

Third, we show that by introducing a modification to the default
activation-based model, namely, by assuming that the accumulation of evidence
for the retrieval of incorrect items is not only slower but noisier, the new model
can provide a fit as good as that of the direct-access model.

The importance of a formal comparison of \posstextcite{LewisVasishth2005}
activation-based model and \posstextcite{McElree2000} direct-access model
lies in disentangling their predictions. Since both models  assume that
dependencies of non-adjacent elements are created via a content-addressable
cue-based retrieval mechanism, they have been used almost interchangeably to
explain interference effects \cite<e.g.>[]{VanDykeMcElree2006}. For
experiments that draw inferences from differences in means, these two models
yield  identical predictions for the inhibitory effect of similarity-based
interference: namely, longer reading times at the retrieval of a dependent
and/or a reduction of comprehension accuracy when several items share a feature
associated with a retrieval cue. However, these models are based on different
underlying assumptions. The activation-based model follows ACT-R assumptions
\cite{AndersonEtAl2004}; in this framework, the activation of the items in
memory determines the retrieval accuracy and latency, and the activation of the
target of retrieval is, in turn, adversely affected by interference. Crucially,
latency and accuracy are not deterministic because activation fluctuates due to
noise in the system. We show that this process can be seen as a lognormal race
between accumulators of evidence with a single variance for all the
accumulators, where activation represents the rate of accumulation of evidence.
In contrast, the direct-access model assumes a model of memory where only the
probability of retrieval can be affected by interference, while items take the
same time to be retrieved (if they are not in the focus of attention as it is
the case for non-local dependencies). In this model, differences in latencies
are a by-product of the possibility of backtracking and repairing incorrect
retrievals. These different assumptions lead to a different behavior in the
relationship between reading times and response accuracy on a trial-level basis
which cannot be examined by only comparing mean reading times or accuracy
between conditions. While acceptability judgment tasks with speed-accuracy
trade-off (SAT) allow a finer grain look at the reaction times and have been
used to argue in favor of the direct-access model \cite<see, for
example,>[]{VanDykeMcElree2011}, until now there has been  no computational
evaluation of the model in reading for comprehension.

While the activation-based model uses the declarative retrieval module
of ACT-R, which has been shown to be an empirically successful model 
\cite<e.g.,>[]{AndersonEtAl1998,AndersonReder1999,VanRijnAnderson2003},
our findings show that its default implementation cannot account for incorrect
retrievals that were generally faster than the correct ones in our data. The
model cannot account for this pattern because items in memory that match the
retrieval cues will have  higher activation on average than competitors that
match  the retrieval cues only partially. The higher activation on average
leads in turn to overall faster retrievals. In contrast, the direct-access model can successfully accommodate faster incorrect retrievals. This is done
by assuming that reading times associated with correct responses are generated
from a mixture distribution of fast directly accessed correct retrievals at
the first attempt together with slower backtracked and repaired retrievals. Reading times
associated with incorrect responses, in contrast, belong to a faster
distribution of retrieval latencies of items that are directly accessed. It
should be noted that this repair mechanism that explains slow correct
retrievals could in principle be added to the activation-based model, but it
would lead to an unidentifiable model. The direct-access model, however, is able
to account for the data with a very simple architecture that can integrate
this repair mechanism. 

While the simple architecture of the direct-access model may be preferred on
grounds of parsimony,  the activation-based model has some arguably desired
characteristics: it is compatible with a sequential sampling framework
(\citeNP<such as the drift diffusion model:>[]{Ratcliff1978}; \citeNP<the leaky
competitive accumulator:>[]{UsherMcClelland2001}; \citeNP<linear deterministic
models:>[among others]{HeathcoteLove2012}). In the sequential sampling framework, decisions (such as which
is the right dependent that needs to be retrieved) are considered a process of
noisy accumulation of evidence, which has been shown to be compatible with the
behavior of populations of neurons
\cite<e.g.,>[]{ZandbeltEtAl2014}. In addition, sequential sampling has
been also linked to  theories of optimality
\cite{RatcliffEtAl2016,SummerfieldTsetsos2015}, which compare how an
ideal agent would perform (given the levels of uncertainty in the stimuli) 
 with the actual behavior of participants. 

The sequential sampling framework could still be useful to explain
retrieval, if we assume that the retrieval process behaves similarly to
other more complex accumulator models such as the linear ballistic accumulator
\cite{BrownHeathcote2008}{, since, as explained earlier, 
they can account for fast and slow errors \cite{WagenmakersEtAl2008}. }
However, these models lose the close connection with the ACT-R framework that
motivated the lognormal race, which underlies the activation-based model. In
addition, given that these models are more complex than
the lognormal race model, { it may be the case that they overfit the
data} and it is not clear whether their fit would be comparable to the fit of the
direct-access model. A potential future direction of this work would be to
evaluate different plausible accumulator models as models of retrieval.

We also relaxed one of the assumptions of ACT-R to capture
the patterns of the data: Here we assumed that the activation of chunks that
match  the retrieval cues only partially is not only lower but also noisier.
This amounts into assuming different variances for the different
accumulators. \citeA{HeathcoteLove2012} show  that when the accumulators
associated with incorrect responses have a larger variance than the
accumulator of correct responses, the model can account for fast errors on
average. For simplicity, we assumed one variance for the accumulator of the
target, and one for the competitors and failure accumulators. While our study
shows that this is enough to account for the pattern in the data, nothing would
prevent all accumulators from having different variances.

\begin{lrbox}{\LstBox}
\begin{lstlisting}[basicstyle=\linespread{0.8}\footnotesize]
install.packages(c("dplyr", "tidyr", "ggplot2","cowplot","shiny")) #if needed
library(shiny) #load shiny
runUrl("http://www.ling.uni-potsdam.de/~nicenboim/code/race-plot.zip") 
\end{lstlisting}
\end{lrbox}

\label{p:repl}{Both the activation-based model  with different variances and 
 the direct-access model showed an equally good fit to the data. In order to
 investigate these models' relative fit, future work should replicate the
 classical interference results
 \cite<e.g.,>[]{VanDykeMcElree2006,vanDyke2007,VanDykeMcElree2011},
 while including reading times and questions probing the comprehension of the
 relevant dependencies.} { {Given that the posteriors
 for the interference effect in the relevant parameters had a broad distribution
 and were not too informative, a large-sample replication of the classical
 interference experiments would allow us to verify the main predictions of the
 cue-based retrieval models.}}

There are other phenomena that the models could explain. These are:
(i) the facilitatory
interference effects found in ungrammatical sentences
\cite{WagersEtAl2009}, (ii) the ambiguity advantage in
relative clauses \cite{TraxlerEtAl1998} and the effect of task demands
\cite{SwetsEtAl2008}, and (iii) good-enough processing
\cite{FerreiraEtAl2002}.  Since some of the predictions
of the activation-based model with different variances are not very intuitive, we
provide an \emph{R} script called \emph{race-plot} using the  \emph{Shiny} package
\cite{shiny} that can help visualizing the predictions.\footnote{The
application can be accessed in the browser with the following commands in R:

\usebox{\LstBox}
}

\paragraph*{Facilitatory interference}
\citeA{WagersEtAl2009} noticed that the so-called  number attraction effect
in ungrammatical sentences such as \eqref{ex:agreement}, that is,  the speedup in
\emph{are} in \eqref{ex:ungramat} vs.\ \eqref{ex:ungramnoat}, could be
accounted by \posstextcite{LewisVasishth2005} activation-based model.

\begin{exe}
\ex   \label{ex:agreement}
\begin{xlist}
\ex[*] {The key$_{+sing}$ to the cabinet$_{+sing}$ are in the box. } \label{ex:ungramnoat}
\ex[*] {The key$_{+sing}$ to the cabinets$_{+plur}$ are in the box. } \label{ex:ungramat}
\end{xlist}
\end{exe}

 In sentences like \eqref{ex:agreement}, a cue-based retrieval mechanism would
assume that a retrieval is initiated at the  verb (\emph{are}) with at least
two retrieval cues:  grammatical subject and plural. In sentence
\eqref{ex:ungramnoat}, \emph{the key} matches one of the retrieval cues,
because it is the grammatical subject, but mismatches the plural cue. In
sentence \eqref{ex:ungramat}, both nouns partially match the retrieval cues:
\emph{the key} matches the grammatical subject cue, while
\emph{the cabinets} matches the plural cue. An interesting prediction of the
activation-based model \cite<confirmed by  experimental findings;
see>[for a meta-analysis]{JaegerEtAl2017} is that reading times are
\textit{faster} at the verb in \eqref{ex:ungramat} than in
\eqref{ex:ungramnoat}. This is so because a situation with no unambiguous
match (both nouns are partial matches) leads to statistical facilitation
\cite{Raab1962}, that is, an overall speedup when we examine mean reading
times (facilitatory interference).  For facilitatory interference in
ungrammatical sentences, the predictions of the default activation-based model
and the activation-based model with different variances are the same. This
situation can be simulated using \emph{race-plot} script mentioned before, by
assigning arbitrary (but plausible) activations to the candidates to retrieval
in \eqref{ex:ungramnoat} and \eqref{ex:ungramat}: In  \eqref{ex:ungramnoat},
\emph{the key} (partial match) can be assigned an activation of 4 and
\emph{the cabinet} (no match) an activation of 2.5  (and  $\sigma=1.5$); this
would result in a mean reading time of approximately 832 ms. Notice that since
the process is not deterministic, different simulations will show different
retrieval times; the relationship between the conditions, however, should
hold on average. In
\eqref{ex:ungramat}, \emph{the key} (partial match) can be assigned an
activation of 4 and \emph{the cabinets} (partial match) an activation of 3.5
(since they will not necessarily reach exactly the same activation); this
would result in a faster reading time on average,  approximately 692 ms.

In contrast to the activation-based model, the direct-access model would not
predict a difference in reading times at the verb between
\eqref{ex:ungramnoat} and \eqref{ex:ungramat}. This is the case since
increased reading times depend only on backtracking, which would only occur to
repair an initially incorrect retrieval. In ungrammatical sentences with
partial match, it is unclear how the repair would work, and why there would be
more backtracking in
\eqref{ex:ungramnoat} than in \eqref{ex:ungramat}.

The predictions of the activation-based model, however, have not been
investigated taking into account both reading times and comprehension.  Even
if a speedup compatible with facilitatory interference has been reported in
the literature, the activation-based model would be accounting for
facilitatory interference only if participants reach  a different
interpretation of the sentence in \eqref{ex:ungramat} more often than in
\eqref{ex:ungramnoat}.\footnote{Notice that even though it is unlikely that
readers would understand that \emph{the cabinets are in the box}, it may be
that the sentence is reanalyzed when the parser reaches \emph{box}.}

\paragraph*{The ambiguity advantage in relative clauses and task-demands effects}
The so-called ambiguity advantage is based on the observation of
\citeA{TraxlerEtAl1998}, who found a speedup at \emph{mustache} in
ambiguous conditions such as \eqref{ex:global} in comparison with unambiguous
conditions such as \eqref{ex:high} and \eqref{ex:low}, where \emph{mustache} is
the disambiguating word.

\begin{exe}
\ex   \label{ex:ambig}
\begin{xlist}
\ex The driver of the car that had the mustache was pretty cool. (high attachment) \label{ex:high}
\ex The car of the driver that had the mustache was pretty cool. (low attachment) \label{ex:low}
\ex The son of the driver that had the mustache was pretty cool. (globally ambiguous) \label{ex:global}
\end{xlist}
\end{exe}

The account of the activation-based model with different variances is very
similar to the unrestricted race model proposed by
\citeA{vanGompelEtAl2000}, which predicts statistical facilitation in the
case of ambiguity. According to the unrestricted race model,  the parser
starts building all possible structures simultaneously. While the time taken
depends on plausibility, it is also affected by noise. This means that the
adopted structure in each trial is the one that takes the least time, leading
to shorter time on average when there are more candidates.

The activation-based model with different variances would yield similar
predictions to the unrestricted race model if the inhibitory effect of
interference in \eqref{ex:global} is sufficiently small. Given the relatively small
magnitude of the interference effect in the literature
\cite{JaegerEtAl2017,NicenboimEtAl2016NIG}, this is likely to be the
case. In unambiguous cases such as \eqref{ex:high} and \eqref{ex:low}, there
is only one NP that matches  the retrieval cue: ``being capable of having a
mustache'' (i.e., \emph{the driver}). In ambiguous cases such as \eqref{ex:global},   there are two
NPs that match the retrieval cue  (i.e, \emph{The son} and \emph{the
driver}). Therefore, we would expect statistical facilitation (similarly to
the case of facilitatory interference) leading to faster reading times on
average. This situation can be simulated using the \emph{race-plot} script similarly as
before: In  \eqref{ex:high} or \eqref{ex:low}, \emph{the driver} (full match)
can be assigned an activation of 5 (and $\sigma = 1$), and \emph{the car}
(partial match) can be assigned an activation of 2.5 (and  $\sigma = 2$); this
would result in a mean reading time of approximately 416 ms. In
\eqref{ex:global}, both \emph{The son} and \emph{the driver} should have
similar activation since there is no penalty component involved, both are a
full match. However, the cue ``can have a mustache'' does not uniquely identify any
candidate. Given the small magnitude of inhibitory interference effects, we
could assume an activation of 4.8 (instead of 5) and the same variance since
there is no mismatch (i.e. $\sigma = 1$) for both NPs. This would result in a
faster reading time on average, approximately 349 ms.

Furthermore, the activation-based model may be able to account for
\posstextcite{LogacevMultiple} observation that the parser seems to behave in
a way that resembles a race between interpretations (low attachment vs.\ high
attachment) but it is also task-dependent \cite<as assumed
by>[]{SwetsEtAl2008}.
This could be achieved by setting the timeout (the parameters of the accumulator
associated with the retrieval failure) to be task-dependent: longer timeouts
when instructions or context encourage attentive reading and shorter timeouts
when a full interpretation is not needed for successfully completing  the
experimental task.

In this case, the direct-access model could also predict the ambiguity
advantage in a very simple way: While in \eqref{ex:high} or \eqref{ex:low} it
is possible to retrieve the incorrect NP (i.e. ``the car'') leading to a
certain proportion of slower backtracked retrievals, in  \eqref{ex:global}
there should only be fast directly accessed retrievals, since both NPs (i.e.
``The son'' and ``the driver'') are correct targets. In addition, for the
direct-access model, the proportion of incorrect retrievals that are
backtracked could be task dependent, with a larger proportion of backtracking
associated with deeper processing. However, the predictions of both models would not be
identical. The direct-access model predicts that the reading times at the
disambiguating region when the incorrect interpretation (or no interpretation)
is held in  \eqref{ex:high} or \eqref{ex:low} would be identical to the
reading times in \eqref{ex:global}. In contrast, for the activation
based-model with different variances, the relationship between reading times
at the different conditions would depend on comprehension accuracy. Future
work that includes measures of reading times and queries for the comprehension
of the relative clause, as well as manipulates task demands could compare the
activation-based model with different variances, the direct-access model, and
the model presented in \citeA{LogacevMultiple}, which subsumes the
unrestricted race model and allows it to be task-dependent.

\paragraph*{Good-enough processing}
While a comprehensive alternative to good-enough processing is out of the
scope of this section \cite<see>[for a complete
overview]{Christianson2016}, it should be noticed that without further
assumptions  the activation-based model with different variances and the
direct-access model can account for manipulations that show (sometimes
unexpectedly) fast reading times which have been attributed to good-enough
processing. For the activation-based model with different variances, this can
be achieved by associating the timeout with either task demands as suggested
previously or also with individual differences. An increase of either timeout
speed (i.e., the rate of accumulation of the retrieval failure) or an increase
of its noise (i.e., the variance of the accumulator associated with retrieval
failure) would lead to more frequent shallow parses with incomplete
dependencies which are read faster. Thus, experiments that  probe  the
comprehension of certain dependencies less often may lead to faster (and maybe noisier) timeouts, which would in turn lead to shorter mean reading
times. Individual differences in participants such as working memory capacity
may have a similar effect, with lower capacity leading to faster and noisier
timeouts in the retrieval process.

Similarly for the direct-access model, the probability of backtracking could be
affected by  task demands  or by individual differences: A less demanding task
would reduce reading times  and  comprehension accuracy on average by
discouraging backtracking. Individual differences may have a similar effect,
participants with lower working memory capacity may be less prone to
backtracking. As suggested before, this could be assessed in future work by
including measures of reading times and comprehension accuracy of the relevant
dependencies.

\section{Conclusion}

We have provided an evaluation of two theoretically grounded and empirically
successful models in explaining similarity-based interference in sentence
comprehension: \posstextcite{LewisVasishth2005} activation-based model built 
under the assumptions of ACT-R \cite{AndersonEtAl2004}  and
\posstextcite{McElree2000}   direct-access model.  We also
evaluated a third model, a variation of the activation-based model.

Our evaluation, which consisted in implementing these models in a Bayesian
hierarchical framework, confirms that, as expected,  both the activation-based
and direct-access  models can in principle explain interference effects.
However, posterior predictive checks and model comparison using
cross-validation show  that some aspects of the data fit better under the
direct-access model, in particular, the default activation-based cannot
predict that, on average, incorrect retrievals would be faster than correct
ones.

Finally, we show that by introducing a modification of the activation model,
namely, by assuming that the accumulation of evidence for the retrieval of
incorrect items is not only slower but noisier (i.e., different variances for
the correct and incorrect items), the new model can provide a fit as good as
that of the direct-access model.

\bibliography{phdlibbibtex}


\appendix

\section{Implementation of the activation-based model in Stan} \label{app:ABM}

The Stan code (shown in Listing \ref{ls:ABM}) was fit to a Latin-squared
design, where only  the sentences of the original experiment
\cite{NicenboimEtAl2016NIG} with  questions that queried the subject of
the embedded verb was kept, and it used a non-centered parameterization to
improve convergence
\cite<for details see: >[]{PapaspiliopoulosEtAl2007,StanManual2.9} in Stan
\cite{Stan2016}. However, to improve clarity, we ignore  that each
participant did not respond to each experimental item, and we assume a centered
parametrization in the equations below.

Let $i = 1,..., N_{subj}$, $j = 1,... , N_{items}$, and $c =
1,...,N_{choices}$  index participants,  items,  and choices in the
multiple-choice questions (1 is the correct response, the target of the
retrieval, 2 and 3 are incorrect responses, the competitors, and 4 is the
option ``I don't know'', which represents a failed retrieval) respectively.
Let $w_{i,j}$, and $RT_{i,j}$ denote the response selected and the reading
times at the auxiliary verb (\emph{hatte}) for  subject $i$ to the item $j$. 
Then we assume that reading times have the following distribution:

\begin{equation} \label{eq:sampling}
RT_{i,j}  \sim \psi_i +  lognormal(b - \alpha_{i,j,c=w} ,\sigma)
\end{equation}

where $\psi_i$ is a by-subject shift, $b$ is an arbitrary threshold
(set to 10), and $\alpha_{i,j,c=w}$ represents the rate of accumulation of
the ``winner'' accumulator. The rest of the accumulators that did not win the
race must have been slower in that specific trial. From this it follows that the 
accumulators that lost the race have a potential $RT_{i,j,\forall c,c \neq w}$ which
is larger than the observed value $RT_{i,j}$.

If all the answers are selected at least once (and if not, we can safely
remove the accumulator since its rate of accumulation is so low that it never
wins), the race turns into a problem of censored data, where the reading
times, $RT_{i,j,\forall c,c \neq w}$, below a lower bound, $RT_{i,j,c = w}$, never
occur. In order to calculate the posterior of the rate of accumulation,
$\alpha$, of all the accumulators, we cannot ignore the censored data
\cite<pp. 224-227>{GelmanEtAl2014}. However, it is not necessary to
impute values, and the values can be integrated out
(\citeNP<>[pp. 107--110]{StanManual2.9}; \citeNP<>[pp. 224-227]{GelmanEtAl2014}).
Each censored data point has a probability of

\begin{align} \label{eq:prob}
Pr[RT_{n,\forall c, c \neq w}  > RT_{n,c=w} ] &= \int_{RT_{n,c=w}}^{\infty} lognormal(RT_{n,\forall c, c \neq w} - \psi | b- \alpha_{\forall c,c \neq w}  ,\sigma) \cdot dRT_{n,\forall c, c \neq w} \\
                                          &= 1-\Phi \left( \frac{log(RT_{n,c=w} - \psi) -  (b - \alpha_{\forall c,c \neq w}) }{\sigma} \right) 
\end{align}

where $\Phi()$ is the cumulative distribution function of the standard normal
distribution. Since the shifts of the distribution, $\psi_i$, must be positive, to ensure
convergence of the model we exponentiate a term that is associated with a
general shift of the whole reading times distribution, $\psi'$, and a term
that represents the by-participants adjustment, $\psi'_i$:

\begin{equation}
\psi_i = exp(\psi' + \psi'_i) \label{eq:psi}
\end{equation}

with the following priors for the by-participant component:

\begin{align}
\psi'_i &\sim normal(0,\tau_\psi)  \label{eq:psii} \\  
\tau_\psi &\sim normal(0,10); \text{ with }\tau_\psi>0  \label{eq:taupsi} 
\end{align}

In addition,  each  $\psi_i$  must be smaller than the shortest reading time
of each participant  $i$ (recall that the shift is the lower bound of the
distribution). We satisfied the constraint on the upper bound with the
following prior on the general shift. { {We show in
Eq.~\eqref{eq:psi-tr} the parameter $\psi'$ back-transformed from unit-scale;
notice that the standard deviation of the normal prior is in log-scale
(multiplied by ten to have a weaker prior), because this parameter is
exponentiated in \eqref{eq:taupsi} and forms the shift of the distribution.}}

\begin{align} 
{\psi' \sim normal(0,10 \cdot log(mean(RT))); \text{ with } \psi' < U }\label{eq:psi-tr}
\end{align}

\noindent
where a normal distribution is truncated on the upper limit, $U$, which is the
smallest difference between $log(RT)$ and $\psi'_i$.

We assume that the rates of accumulation depend on the experimental condition
(high or low interference) and that the rates may be affected by participants
and by items. We can express this in matrix notation for each accumulator as
follows:

\begin{equation} \label{eq:alpha-fe-re}
\mathbf{\alpha}_{c} = \mathbf{X} \boldsymbol{\beta}_{c} + \mathbf{X} \mathbf{u}_{c}  + \mathbf{X} \mathbf{v}_{c}
\end{equation}

Here  $\mathbf{X}$ is the $N_{obs}
\times N_{pars}$ model matrix, with the number of parameters (so-called fixed
effects), $N_{pars}$, being two: intercept and condition. Each
$\boldsymbol{\beta}_{c}$ is a vector of length $N_{pars}$ with the estimates
of the  fixed-effect parameters for the accumulator associated with the choice
$c$. Each $\mathbf{u}_{c}$ and $\mathbf{v}_{c}$ are the by-participants and
by-item adjustments to the fixed effects estimates (so-called random-effects)
for the accumulator $c$. We used weakly informative priors for all the
parameters (some estimates were reparametrized in the Stan implementation, see
Listing \ref{ls:ABM} for details).

Here, $\beta_{0,c}$ are the intercepts of the fixed effects for choice $c$,
$log(mean(RT_c))$ is  the logarithm of the mean of the reading times when
option $c$ was selected, and $\beta_{1,c}$ represents the slopes of the
fixed effects (i.e., the effect of interference).

All the random-effects, $\mathbf{u}_{c}$, and $\mathbf{v}_{c}$, were assumed
to be sampled from two multivariate normal distributions with means of zero.
The prior of the standard deviations of the random effects was $normal(0,10)$.
We placed lkj priors on the random effects correlation matrices with shape
parameter $\eta=2$ \cite<see>[]{LewandowskiEtAl2009,SorensenEtAl2016}.

 \singlespacing
\begin{lstlisting}[label={ls:ABM},numbers=left,frame=single,
caption={Stan code for the activation-based model},escapechar=@]
functions {
    vector mean_of_X_by_Y(vector X, int[] Y){
    int N_rows;
    int N_groups;
    N_rows = num_elements(X);
    if(N_rows!= num_elements(Y)) 
      reject("X and Y don't have the same length");
    N_groups = max(Y);
    { # matrix with a column for each group of Y, 
      #and 1 if Y belong to the group:
      matrix[N_rows, N_groups] matrix_1; 
      for(r in 1:N_rows)
        for(g in 1:N_groups)
        matrix_1[r,g] =  (g ==  Y[r]);  
      return(((X' * matrix_1)   // sum of Xs per group
      // divided by matrix_1^T * matrix_1 (number of times each group appears):
      / crossprod(matrix_1))');     
    }
  }
  real psi_max(vector u_psi, int[] subj, vector RT) { 
    // This function ensures that psi is correctly truncated on the correct 
    // upper limit, U in Eq. @\eqref{eq:psi-tr}@ 
    real psi_max;
    psi_max = positive_infinity();
    for (i in 1:num_elements(RT))
         psi_max = fmin(psi_max, log(RT[i]) - u_psi[subj[i]]);
    return (psi_max);
  }
  real race(int winner, real RT, row_vector alpha, real b, real sigma, real psi)
   {
    real log_lik;
    int N_choices;
    real rate;
    real shifted_RT;

    N_choices = cols(alpha);
    shifted_RT = RT - psi;

    log_lik = 0;
    for(c in 1:N_choices){
      if(c == winner){    
    // loglik due to distribution of observed RTs; Eq. @\eqref{eq:sampling}@ 
        log_lik = log_lik + lognormal_lpdf(shifted_RT|b - alpha[c], sigma);
      } else {
    // loglik due to censored RTs of loser candidates; Eq. @\eqref{eq:prob}@ 
        log_lik = log_lik + lognormal_lccdf(shifted_RT|b - alpha[c], sigma);
      }
    }
    return(log_lik);
    }
}
data { 
  int<lower = 0> N_obs; 
  int<lower=1> N_subj;                 
  int<lower=1> N_item;       
  int<lower = 1> N_choices; 
  int<lower = 1, upper = N_choices> winner[N_obs]; // response selected
  vector<lower = 0>[N_obs] RT;
  // for fixed effects
  int N_coef;  //intercept + predictors 
  matrix[N_obs,N_coef] x; 
  // for random effects by subject:
  int<lower=0> N_coef_u;  
  int<lower=1> subj[N_obs];
  matrix[N_obs,N_coef_u] x_u;
  // for random effects by items:
  int<lower=0> N_coef_w;  
  int<lower=1> item[N_obs];  
  matrix[N_obs,N_coef_w] x_w;

}
transformed data {
  real b; 
  real min_RT;
  real logmean_RT;
  row_vector[N_choices] logmean_RT_w;
  matrix[N_obs,N_coef-1] x_betas;
  int N_tau_u;
  int N_tau_w;

  b = 10;
  min_RT = min(RT);
  logmean_RT = log(mean(RT));
  logmean_RT_w = (log(mean_of_X_by_Y(RT, winner)))';
  x_betas = x[,2:N_coef]; // x without intercept
  N_tau_u = N_coef_u * N_choices; 
  N_tau_w = N_coef_w * N_choices;
}
parameters{
  row_vector[N_choices] beta_0raw; 
  vector<lower = 0> [N_tau_u]  tau_u;     
  cholesky_factor_corr[N_tau_u] L_u;  
  matrix[N_tau_u, N_subj] z_u;
  vector<lower = 0> [N_tau_w]  tau_w;     
  cholesky_factor_corr[N_tau_w] L_w;  
  matrix[N_tau_w, N_item] z_w;
  real<lower = 0> sigma;
  vector[N_subj] u_psi;
  real<lower = 0> tau_psi; 
  real<upper = psi_max(u_psi, subj, RT) / logmean_RT> psi_p_raw;
  matrix[N_coef-1,N_choices] beta;
}
transformed parameters {
  real psi_p;
  vector[N_obs] psi;
  matrix[N_coef_u,N_choices]  u[N_subj];         
  matrix[N_coef_w,N_choices]  w[N_item];    
  matrix[N_obs, N_choices] alpha; 
  row_vector[N_choices] beta_0; 

  {  
    matrix[N_tau_u, N_subj] u_long; 
    matrix[N_tau_w, N_item] w_long; 
    // Optimization through Cholesky Fact:
    u_long = diag_pre_multiply(tau_u, L_u) //matrix[N_choices,N_choices]
        * z_u;  
    w_long = diag_pre_multiply(tau_w, L_w) //matrix[N_choices,N_choices]
        * z_w;
    for (i in 1:N_subj)
       u[i] = to_matrix(u_long[,i],N_coef_u,N_choices,0);
    for (j in 1:N_item)
       w[j] = to_matrix(w_long[,j],N_coef_w,N_choices,0);
  }
  // Unit-scaling:
  beta_0 = b - beta_0raw .* logmean_RT_w;  
  psi_p = psi_p_raw * logmean_RT;   
  // Fixed effects in alpha; Eq. @\eqref{eq:alpha-fe-re}@:
  alpha = rep_matrix(beta_0, N_obs) + x_betas * beta;
  for (n in 1:N_obs) {
    // Shift adjustment for every observation; Eq. @\eqref{eq:alpha-fe-re}@:
    psi[n] = exp(psi_p + u_psi[subj[n]]);
    // Add random effects
    for (uu in 1:N_coef_u)
      alpha[n] = alpha[n] + x_u[n, uu] * u[subj[n], uu];
    for (ww in 1:N_coef_w)
      alpha[n] = alpha[n] + x_w[n, ww] * w[item[n], ww];
  }
}
model {
  vector[N_obs] log_lik;
  beta_0raw ~ normal(0, 10);
  to_vector(beta) ~ normal(0, 10);  
  tau_u ~ normal(0, 10);
  L_u ~ lkj_corr_cholesky(2.0);
  to_vector(z_u) ~ normal(0, 10); 
  tau_w ~ normal(0, 10);
  L_w ~ lkj_corr_cholesky(2.0);
  to_vector(z_w) ~ normal(0, 1); 
  sigma ~ normal(0, 10);
  psi_p_raw ~ normal(0, 10);
  u_psi ~ normal(0, tau_psi); // Eq. @\eqref{eq:psii}@
  tau_psi ~ normal(0, 10); // Eq. @\eqref{eq:taupsi}@
  for (n in 1:N_obs)
      log_lik[n] = race(winner[n], RT[n], alpha[n], b, sigma, psi[n]);
  target += log_lik;
}
\end{lstlisting}

\section{Implementation of the direct-access model in Stan}\label{app:DAM}

The  code (shown in Listing \ref{ls:DAM}) was fit to  the same data as
the activation-based model. As before, to improve clarity, we ignore  that
each subject did not respond to each item and we assume a centered
parametrization.

Let $i = 1,..., N_{subj}$, $j = 1,... , N_{items}$, and $c = 1,...,N_{choices}$
index participants,  items,  and choices respectively, where choice 1 is the
correct response and choice $N_{choices}$ (which maps to 4) is the response
associated with a retrieval failure. Let $w_{i,j}$, and $RT_{i,j}$ denote the
response selected and the reading times at the auxiliary verb (\emph{hatte})
for subject $i$ to the item $j$.

We implemented the assumptions of the direct-access model, by letting $w$ have
a discrete distribution that follows a one-inflated categorical model, where
additional probability mass is added to the outcome 1 (correct response) due to backtracking with probability $\theta_b$ as follows:

\begin{align}
P(w_n=1 | \boldsymbol{\theta_{i,j}}, \theta_b ) = \theta_{1_{i,j}} + (1-\theta_{1_{i,j}}) \cdot \theta_b \label{eq:dis1}
\\
P(w_n=s |\boldsymbol{\theta_{i,j}}, \theta_b ) = \theta_{s_{i,j}} \cdot (1-\theta_b) \text{, with } s > 1 \label{eq:dis2}
\end{align}

where $\boldsymbol{\theta}$ is a vector of $N_{choices}$ rows that represents
the probability of each option. 

If the answer given is wrong, we assume that there is no backtracking and then
reading times are distributed in the following way:

\begin{equation}
RT_{i,j,\forall w, w>1}  \sim  \psi_i + lognormal(T_{da,i,j},\sigma) \label{eq:errorRT}
\end{equation} 

where $\psi_i$ is a by-subject shift, $T_{da}$ represents the time needed for
the direct access or failure together with extra processes

If the answer given is right, reading times are assumed to have a mixture
distribution. This is so because there are two ``paths'' to reach a correct
response (see Figure~\ref{fig:daprob2}): (i) The chunk that is retrieved is
the correct one (at the first try), and this means that there is direct access
and reading times should belong to a distribution similar to the previous one
as shown in Eq.~\eqref{eq:errorRT}; or (ii) an incorrect chunk (or no chunk) is retrieved but is
backtracked, and this means that  reading times should belong to a
distribution with a larger location than $T_{da,i,j}$, namely,
$T_{da,i,j}+t_{b,i,j}$. Thus RTs should be distributed in the following way:

\begin{align}
RT_{i,j,w=1}  \sim& \psi_i + 
\begin{cases}
 lognormal(T_{da,i,j},\sigma) \text{;  if } y=1 | Categorical( y |\boldsymbol{\theta_{i,j}}) \\ 
 lognormal(T_{da,i,j}+t_{b,i,j},\sigma)  \text{;  if }y \neq 1 \text{ and } z=1 | Categorical(y | \boldsymbol{\theta_{i,j}}) \text{ and } Bernoulli(z | \theta_b)
\end{cases} 
\label{eq:correctRT}
\end{align}

where, from Eq.~\eqref{eq:dis1}, the first component of the mixture
defined in \eqref{eq:correctRT} is the probability of a correct retrieval at the first
attempt conditional on obtaining $one$ as a response ($w_n=1$), and occurs
with probability:

\begin{equation}
 \frac{\theta_{1_{i,j}} }{ \theta_{1_{i,j}} + (1-\theta_{1_{i,j}}) \cdot \theta_b } \label{eq:first}
\end{equation}

and the second component of the mixture \eqref{eq:correctRT} represents the
probability of backtracking when there is an error, this is formulated as  the
probability of an incorrect retrieval in the first attempt conditional on
obtaining $one$ as a response ($w_n=1$):

\begin{equation}
 \frac{1 - \theta_{1_{i,j}} }{ \theta_{1_{i,j}} + (1-\theta_{1_{i,j}}) \cdot \theta_b } \label{eq:repair}
\end{equation}

Furthermore, the categorical model was fit including  a hierarchical structure
in $\boldsymbol{\theta'}$, where $\boldsymbol{\theta'}$ is a vector with
$N_{choices}$ rows with its last row set to zero, so that
$softmax(\boldsymbol{\theta'})=\boldsymbol{\theta}$.\footnote{ The softmax
function is defined as in \citeA{StanManual2.9} by
$softmax(y)=\frac{exp(y)}{\sum_{k=1}^{K} exp(y_k)}$ } This way we ensure the
identifiability of $Categorical(softmax(\boldsymbol{\theta'}))$.

We assume that the probability of each choice depend on the experimental
condition (high or low interference)  and   that the probabilities may be
affected by participants and by items. In matrix notation, the first
$N_{choices}-1$ rows of $\boldsymbol{\theta'}$ are structured as the
activations  in the activation-based model:

\begin{equation}
\theta_{c}' = \mathbf{X} \boldsymbol{\beta}_{c} + \mathbf{X} \mathbf{u}_{c}  + \mathbf{X} \mathbf{v}_{c} \label{eq:theta-fe-re}
\end{equation}

As for the activation-based model, we used weakly informative priors for all
the estimates. The priors for the fixed effects were set with the added
constraint that $\beta_{0,1}$, the intercept of the probability of the correct
choice (the first choice) in logit-space, was constrained to be larger than
$\beta_{0,2..3}$, the intercept associated with the incorrect responses, and
zero (which is the value associated with the last choice):

\begin{align}
\beta_{0,2..3} &\sim normal(0, 10) \label{eq:beta02-3} \\ 
\beta_{0,1} &\sim normal(0, 10) + max(\beta_{0,2..3},0) \label{eq:beta01}\\
\theta_b &\sim beta(1, 1) \label{eq:thetab}
\end{align}

In addition, we assumed a hierarchical structure to the parameters associated
with latencies:

\begin{align}
T_{da,i,j} &= \beta_{0,Tda} + u_{Tda,i} + v_{Tda,j} \label{eq:Tda} \\
T_{b,i,j} &= \beta_{0,Tb} + u_{Tb,i} + v_{Tb,j} \label{eq:Tb} 
\end{align}

with the following priors on the intercepts:
\begin{align}
\beta_{0,Tda} &\sim normal(0, 10 \cdot log(mean(RT))) \label{eq:betatda} \\
\beta_{0,Tb} &\sim normal(0, 10) \label{eq:betatb}
\end{align}

All the random-effects, $\mathbf{u}_{c}$, $u_{tda}$, $u_{tb}$,
$\mathbf{v}_{c}$, $v_{tda}$, $v_{tb}$ were assumed to be sampled from four
multivariate normal distributions  with means of zero: (i) for the subject
adjustment on probabilities of retrieval, (ii) for a similar
adjustment for items, (iii) for the subject adjustment on latencies, and (iv)
for a similar adjustment for items. As before we  placed lkj priors on the
random effects correlation matrices with shape parameter $\eta=2$.

 \singlespacing
\begin{lstlisting}[label={ls:DAM},numbers=left,frame=single,
caption={Stan code for the direct-access model},escapechar=@]
functions {
  real psi_max(vector u_psi, int[] subj, vector RT) {
    real psi_max;
    psi_max = positive_infinity();
    for (i in 1:num_elements(RT))
         psi_max = fmin(psi_max, log(RT[i]) - u_psi[subj[i]]);
    return (psi_max);
  }
  real da(int winner, real RT, row_vector thetap, real theta_b, real T_da, 
               real T_b, real sigma, real psi){
    vector[num_elements(thetap)] thetapT;
    // theta = softmax(thetapT)
    
    real log_P_w1; // log(P(w = 1 | theta, theta_b)); Eq. @\eqref{eq:dis1}@ 
    real log_P_da_gw1; // log(Prob of direct access given winner = 1); Eq. @\eqref{eq:first}@ 
    real log_theta_b_gw1; // log(Prob of backtracking given winner = 1); Eq. @\eqref{eq:repair}@ 

    thetapT = thetap';

    // Eq. @\eqref{eq:dis1}@ in log:
    log_P_w1 = log_sum_exp(categorical_logit_lpmf(1 | thetapT),
                log(theta_b)+ log1m_exp(categorical_logit_lpmf(1|thetapT)));
    // Eq. @\eqref{eq:first}@ in log:
    log_P_da_gw1 = categorical_logit_lpmf(1 | thetapT) - log_P_w1;
    // Eq. @\eqref{eq:repair}@ in log:
    log_theta_b_gw1 = log(theta_b) + 
                    log1m_exp(categorical_logit_lpmf(1 | thetapT)) - log_P_w1;
    if(winner==1) {
    return (log_P_w1 + // Increment on likelihood due to winner=1
                       // Increment on likelihood due to RT:
            log_sum_exp(log_P_da_gw1 + lognormal_lpdf(RT - psi| T_da, sigma),
            log_theta_b_gw1 + lognormal_lpdf(RT - psi | T_da + T_b, sigma) ));
    } else {
      return (log1m(theta_b) +                           // 
              categorical_logit_lpmf(winner | thetapT) + // Eq. @\eqref{eq:dis2}@
              // Increment on likelihood due to RTs; Eq. @\eqref{eq:errorRT}@:
              lognormal_lpdf(RT - psi | T_da, sigma));
    }
  }
}
data {
  int<lower = 0> N_obs; 
  int<lower = 1> N_choices; 
  int<lower = 1,upper = N_choices> winner[N_obs];
  vector<lower = 0>[N_obs] RT;
  int N_coef; 
  matrix[N_obs,N_coef] x;
  int<lower = 0> N_coef_u;  
  int<lower = 1> subj[N_obs];  
  int<lower = 1> N_subj;                 
  matrix[N_obs, N_coef_u] x_u;
  int<lower = 0> N_coef_w;  
  int<lower = 1> item[N_obs];   
  int<lower = 1> N_item;                 
  matrix[N_obs, N_coef_w] x_w; 
}
transformed data {
  real<lower = 0> min_RT;
  real logmean_RT;
  matrix[N_obs, N_coef - 1] x_betas;
  int N_tau_u;
  int N_tau_w;

  min_RT = min(RT);
  logmean_RT = log(mean(RT));
  x_betas = x[, 2:N_coef]; // intercept removed
  N_tau_u = N_coef_u * (N_choices - 1);  
  N_tau_w = N_coef_w * (N_choices - 1);
}
parameters{
  real<lower = 0> sigma;
  real<lower = 0> beta_0_Tdaraw;
  real<lower = 0> beta_0_Tb;
  row_vector[N_choices-2] thetap_incorrect; 
  real<lower = 0> thetap_added;
  matrix[N_coef-1,N_choices-1] beta;
  vector<lower = 0> [N_tau_u]  tau_u;     
  cholesky_factor_corr[N_tau_u] L_u;  
  matrix[N_tau_u, N_subj] z_u;
  vector<lower = 0> [2]  tau_u_RT;     
  cholesky_factor_corr[2] L_u_RT;  
  matrix[2, N_subj] z_u_RT;
  vector<lower = 0> [N_tau_w]  tau_w;     
  cholesky_factor_corr[N_tau_w] L_w;  
  matrix[N_tau_w, N_item] z_w;
  vector<lower = 0> [2]  tau_w_RT;     
  cholesky_factor_corr[2] L_w_RT;  
  matrix[2, N_item] z_w_RT;
  real<lower = 0, upper = 1> theta_b;
  vector[N_subj] u_psi;
  real<lower = 0> tau_psi; 
  real<upper = psi_max(u_psi, subj, RT) / logmean_RT> psi_p_raw;
}
transformed parameters{
  real<lower=0> beta_0_Tda;
  vector[N_obs] T_da;
  vector[N_obs] T_b;
  vector[N_obs] psi;
  matrix[2, N_subj] u_RT; 
  matrix[N_coef_u,N_choices - 1] u[N_subj];
  matrix[2, N_item] w_RT; 
  matrix[N_coef_w,N_choices - 1] w[N_item];
  real psi_p;
  matrix[N_obs, N_choices] thetap; 
  row_vector[N_choices-1] beta_0;

{  
    matrix[N_tau_u, N_subj] u_long; 
    matrix[N_tau_w, N_item] w_long; 
    // Optimization through Cholesky Fact:
    u_RT = diag_pre_multiply(tau_u_RT, L_u_RT) * z_u_RT;   
    w_RT = diag_pre_multiply(tau_w_RT, L_w_RT) * z_w_RT;   
    u_long = diag_pre_multiply(tau_u, L_u) //matrix[N_tau_u,N_tau_u]
        * z_u;  
    w_long = diag_pre_multiply(tau_w, L_w) //matrix[N_tau_w,N_tau_w]
        * z_w;
    for (i in 1:N_subj)
      u[i] = to_matrix(u_long[,i],N_coef_u,(N_choices-1),0);
    for (j in 1:N_item)
      w[j] = to_matrix(w_long[,j],N_coef_w,(N_choices-1),0);
  }

  beta_0[1] = thetap_added + fmax(max(thetap_incorrect), 0); // Eq. @\eqref{eq:beta01}@
  beta_0[2:] = thetap_incorrect; //Eq. @\eqref{eq:beta02-3}@ 
  beta_0_Tda = beta_0_Tdaraw * logmean_RT; // Eq. @\eqref{eq:betatda}@
  psi_p = psi_p_raw * logmean_RT;   
  // Adds so called fixed effects; first summand of Eq. @\eqref{eq:theta-fe-re}@
  thetap[,1:N_choices-1] = rep_matrix(beta_0, N_obs) + x_betas * beta ;
  thetap[,N_choices] = rep_vector(0, N_obs);

  for (n in 1:N_obs) {
    psi[n] = exp(psi_p + u_psi[subj[n]]);
    T_da[n] = beta_0_Tda + u_RT[1,subj[n]] + w_RT[1,item[n]]; // Eq. @\eqref{eq:Tda}@ 
    T_b[n] = beta_0_Tb + u_RT[2,subj[n]] + w_RT[2,item[n]];   // Eq. @\eqref{eq:Tb}@ 
   //adds so called random effects; second two summands of Eq. @\eqref{eq:theta-fe-re}@
    for (uu in 1:N_coef_u)
      thetap[n, 1:N_choices - 1] = thetap[n,1:N_choices - 1] + x_u[n, uu] *
                                u[subj[n], uu];
    for (ww in 1:N_coef_w)
      thetap[n, 1:N_choices - 1] = thetap[n,1:N_choices - 1] + x_w[n, ww] * 
                                w[item[n], ww];
  }
}
model {
  vector[N_obs] log_lik;
  sigma ~ normal(0, 10);
  thetap_added ~ normal(0, 10); // First summand of Eq. @\eqref{eq:beta01}@
  thetap_incorrect ~ normal(0, 10); //Eq. @\eqref{eq:beta02-3}@ 
  to_vector(beta) ~ normal(0, 1);
  psi_p_raw ~ normal(0, 10);
  u_psi ~ normal(0, tau_psi); // Eq. @\eqref{eq:psii}@
  tau_psi ~ normal(0, 10);   //  Eq. @\eqref{eq:taupsi}@
  to_vector(z_u_RT) ~ normal(0, 1); 
  to_vector(z_u) ~ normal(0, 1); 
  tau_u_RT ~ normal(0, 10);
  tau_u ~ normal(0, 10);
  L_u_RT ~ lkj_corr_cholesky(2.0);
  L_u ~ lkj_corr_cholesky(2.0);
  to_vector(z_w_RT) ~ normal(0, 1); 
  to_vector(z_w) ~ normal(0, 1); 
  tau_w_RT ~ normal(0, 10);
  tau_w ~ normal(0, 10);
  L_w_RT ~ lkj_corr_cholesky(2.0);
  L_w ~ lkj_corr_cholesky(2.0);
  theta_b ~ beta(1, 1);  // Eq. @\eqref{eq:thetab}@ 
  beta_0_Tdaraw ~ normal(0, 10);
  beta_0_Tb ~ normal(0, 10);  // Eq. @\eqref{eq:betatb}@
  for (n in 1:N_obs) 
    log_lik[n] = da(winner[n], RT[n], thetap[n], theta_b, T_da[n], T_b[n], sigma,
                 psi[n]);
  target += log_lik;
}
\end{lstlisting}


\section{Recovery of the parameters} \label{app:rec}

We show here the ability of the models to recover the parameters generated from
fake datasets. We used three fake datasets, one for testing each of the models we
discussed in the paper: the activation-based model, the direct-access model, and
the activation-based model with different variances. The procedure for each dataset
and corresponding model is as follows:
\begin{enumerate}
\item We extract the point estimates of a given model. \label{en:point}
\item We generated a dataset assuming that the given model is the \emph{true
generating process}, and the true value of the parameters are the means of the posteriors from \ref{en:point}.
\item We fit the model to its corresponding dataset.
\item We extract the posteriors of the estimated parameters.
\item We show graphically the discrepancy between the estimated posteriors and the
true values.
\end{enumerate} 

\begin{figure}[p]
\begin{knitrout}
\definecolor{shadecolor}{rgb}{0.969, 0.969, 0.969}\color{fgcolor}

{\centering \includegraphics[width=.97\linewidth]{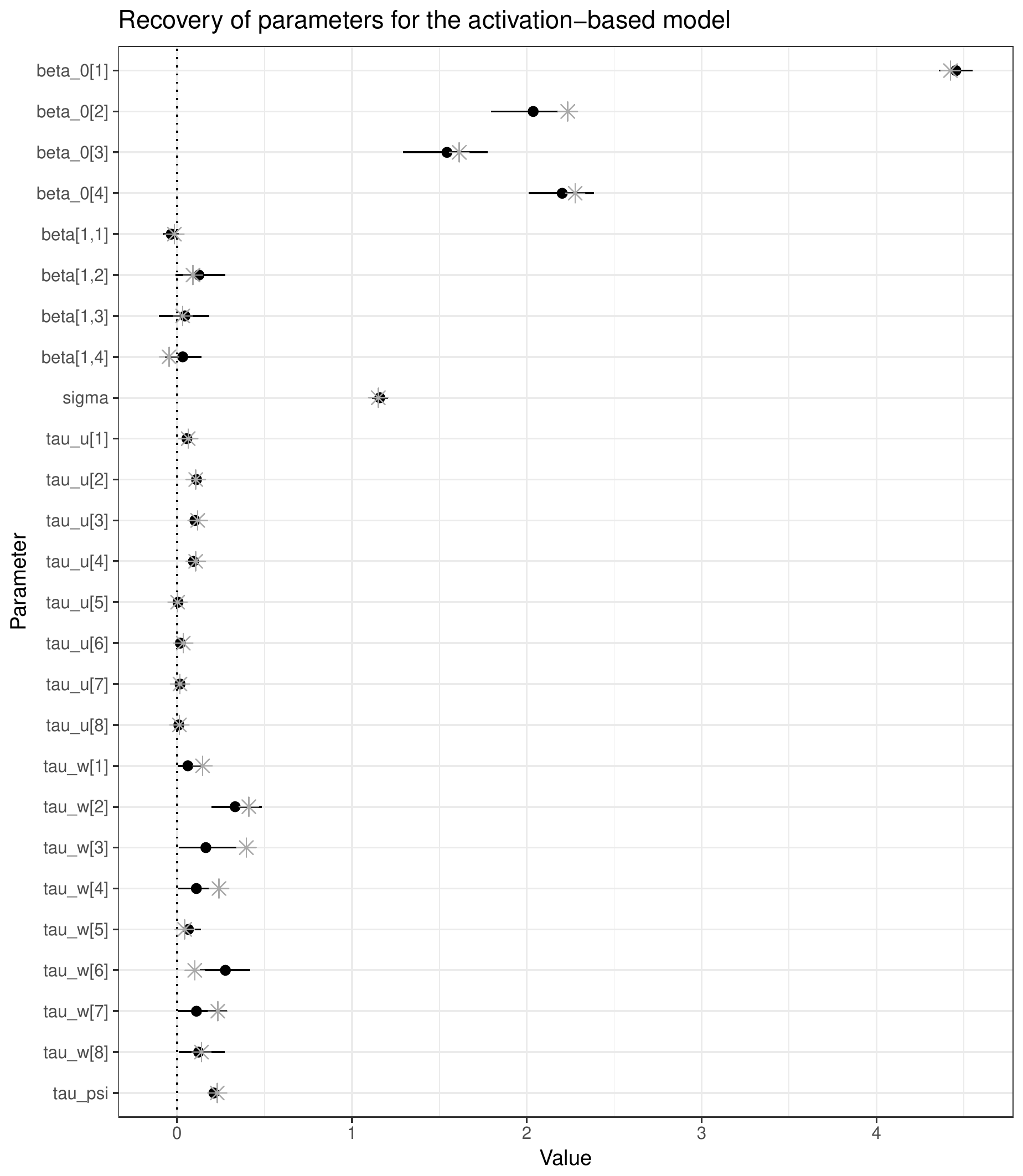} 

}

\end{knitrout}
\caption{Discrepancies between estimated and true value of the parameters of the activation-based model. Black points indicate the mean of the posteriors and the black horizontal lines indicate the 95\% CrIs, the stars indicate the true value of the parameters.} \label{fig:disc-ab}
\end{figure}

\begin{figure}[p]
\begin{knitrout}
\definecolor{shadecolor}{rgb}{0.969, 0.969, 0.969}\color{fgcolor}

{\centering \includegraphics[width=.97\linewidth]{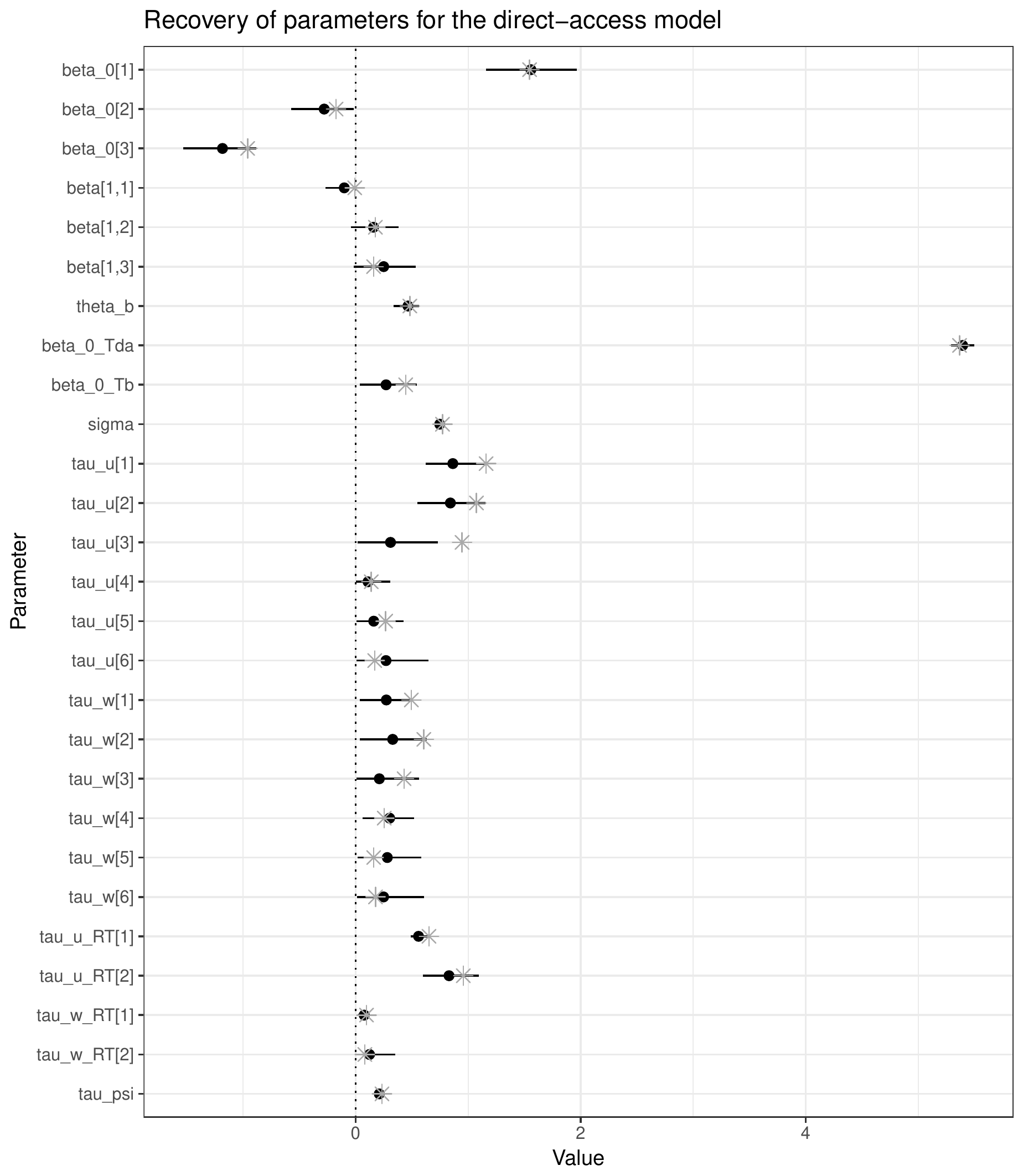} 

}

\end{knitrout}
\caption{Discrepancies between estimated and true value of the parameters of the direct-access model. Black points indicate the mean of the posteriors and the black horizontal lines indicate the 95\% CrIs, the stars indicate the true value of the parameters.} \label{fig:disc-da}
\end{figure}

\begin{figure}[p]
\begin{knitrout}
\definecolor{shadecolor}{rgb}{0.969, 0.969, 0.969}\color{fgcolor}

{\centering \includegraphics[width=.97\linewidth]{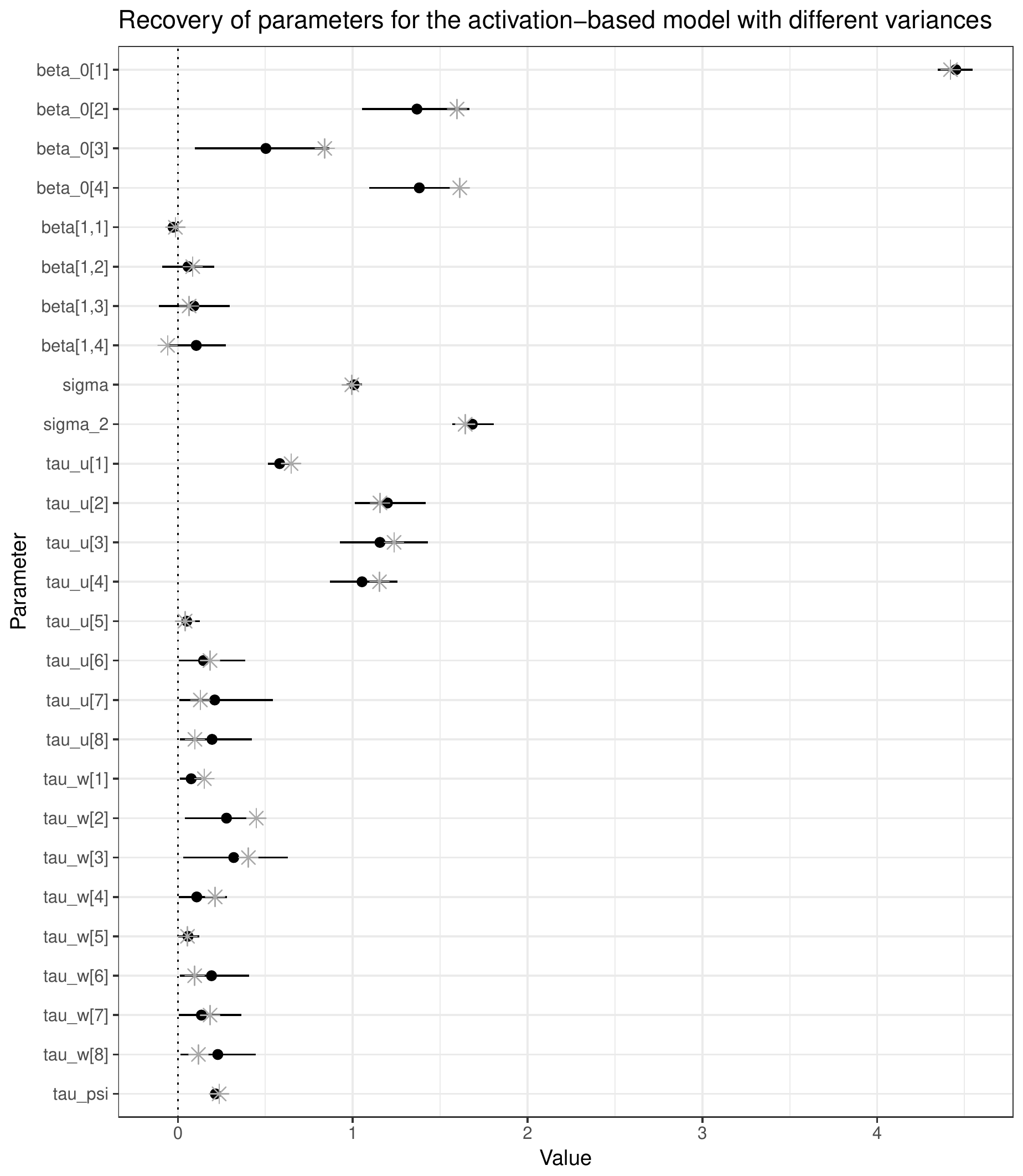} 

}

\end{knitrout}
\caption{Discrepancies between estimated and true value of the parameters of the activation-based model with different variances. Black points indicate the mean of the posteriors and the black horizontal lines indicate the 95\% CrIs, the stars indicate the true value of the parameters.} \label{fig:disc-ab2}
\end{figure}

Figures~\ref{fig:disc-ab}, \ref{fig:disc-da}, and \ref{fig:disc-ab2} show the
true values, and the mean and 95\% CrI of the posterior distribution for the
parameters of the activation-based model, direct-access model, and
activation-based model with different variances respectively.

\end{document}